%% file: HZD_Handbook.tex
\documentclass[onecolumn]{IEEEtran}
\usepackage{graphicx}
\usepackage{amsmath,amssymb,xspace,lscape,epsfig,psfrag,mathrsfs}
\usepackage{latexsym,wrapfig,booktabs,array}
\usepackage{mathabx}
\usepackage{textcomp}
\usepackage{subfig}

\usepackage{hyperref}
\newtheorem{proposition}{Proposition}
\newtheorem{corollary}{Corollary}
\newtheorem{definition}{Definition}
\newtheorem{theorem}{Theorem}

\makeatletter
\DeclareRobustCommand\onedot{\futurelet\@let@token\@onedot}
\def\@onedot{\ifx\@let@token.\else.\null\fi\xspace}

\newcommand{\figcaption}[1]{\def\@captype{figure}\caption{#1}}
\newcommand{\tblcaption}[1]{\def\@captype{table}\caption{#1}}
\makeatother

\def\title#1{{\noindent\Large{\bf #1}\par}}
\def\author#1{\begin{center}{\sc #1\par}\end{center}}

\include{Grizzle_hdr}

\include{Grizzle_HZD_figures}
\begin{document}
\pagestyle{empty}

\title{Virtual Constraints and Hybrid Zero Dynamics for Realizing Underactuated Bipedal Locomotion}
\author{Jessy W Grizzle, University of Michigan, and Christine Chevallereau, CNRS, IRCCyN (Institut de Recherche en Communication et Cybern\'etique de Nantes). }

\section*{Abstract}
Underactuation is ubiquitous in human locomotion and should be ubiquitous in bipedal robotic locomotion as well. This chapter presents a coherent theory for the design of feedback controllers that achieve stable walking gaits in underactuated bipedal robots. Two fundamental tools are introduced, virtual constraints and hybrid zero dynamics. Virtual constraints are relations on the state variables of a mechanical model that are imposed through a time-invariant feedback controller. One of their roles is to synchronize the robot's joints to an internal gait phasing variable. A second role is to induce a low dimensional system, the zero dynamics,  that captures the underactuated aspects of a robot's model, without any approximations. To enhance intuition, the relation between physical constraints and virtual constraints is first established. From here, the hybrid zero dynamics of an underactuated bipedal model is developed, and its fundamental role in the design of asymptotically stable walking motions is established. The chapter includes numerous references to robots on which the highlighted techniques have been implemented.

\section{Introduction}
\label{sec:grizzle:intro}

\input{sections/1-Introduction.tex}

\section{Why Study Underactuation?}
\label{sec:grizzle:why_point_feet}

\input{sections/2-Underactuation.tex}

\section{Hybrid Model of a Bipedal Walker}
\label{sec:grizzle:model}

\input{sections/3-HybridModel.tex}

\section{Physical Constraints and Reduced-order Models}
\label{sec:grizzle:PhysicalConstraints}

\input{sections/4-PhysicalConstraintsJWG.tex}

\section{Virtual Constraints and Zero Dynamics}
\label{sec:grizzle:FdbkDesignApproach}
\input{sections/5-VirtualConstraintsAndZeroDynamics_v02.tex}

\section{Stability Analysis with Hybrid Zero Dynamics}
\label{sec:grizzle:AnalysisZeroDynamics}

\input{sections/6-AnalysisZeroDynamics_v03JWG.tex}

\section{Design of the Virtual Constraints}
\label{sec:grizzle:DesignZeroDynamics}
%
\input{sections/7-DesignZeroDynamics.tex}

\section{Implementations on Real Robots}
\label{sec:grizzle:exp}
\input{sections/8-ExperimentalImplementation.tex}

\section{Further results and Open questions }
\label{sec:conclusion:Summary}
\input{Challenges.tex}


\section{Acknowledgements}
The work of J.W. Grizzle has been generously supported by NSF grants  EECS-1525006, ECCS-1343720 and CNS-1239037. The work of C. Chevallereau is supported by ANR Equipex Robotex project.

\bibliographystyle{plain}
\bibliography{Grizzle_database}


\end{document}

%% file: Grizzle_hdr.tex
\newcommand{\aq}{q_{\rm c}}
\newcommand{\daq}{\dot{q}_{\rm c}}
\newcommand{\ddaq}{\ddot{q}_{\rm c}}
\newcommand{\ulbl}{\rm f}
\newcommand{\uq}{q_{\ulbl}}
\newcommand{\duq}{\dot{q}_{\ulbl}}
\newcommand{\dduq}{\ddot{q}_{\ulbl}}
\newcommand{\usigma}{\sigma_{\ulbl}}
\newcommand{\dusigma}{\dot{\sigma}_{\ulbl}}

\newcommand{\zero}{{\rm zero}}

\newcommand{\qconstrained}{q_{\rm c}}
\newcommand{\qfree}{q_{\rm f}}

\newcommand{\dqfree}{\dot{q}_{\rm f}}

\newcommand{\ddqfree}{\ddot{q}_{\rm f}}

\newcommand {\eqn}     [1] {(\ref{eqn:#1})}

\newcommand {\Fig}     [1] {Figure~\ref{fig:#1}}
\newcommand {\fig}     [1] {Fig.~\ref{fig:#1}}

\newcommand{\real}       [1] { {{\mathbb R}^{#1}} }

\newcommand{\Bezier} [0] {B\'ezier}
\newcommand{\Poincare}[0] {Poincar\'e}

\newcommand{\numlinks}{N}
\newcommand{\Q}{{\cal Q}}
\newcommand{\X}{{\cal X}} 

\newcommand{\KE}{K}
\newcommand{\PE}{V}
\newcommand{\pswingfootv}{p^{\rm z}_2}
\newcommand{\vswingfootv}{\dot{p}^{\rm z}_2}

\newcommand{\impactmap}{\varDelta}
\newcommand{\impactmappos}{\varDelta_{q}}
\newcommand{\impactmapvel}{\varDelta_{\dot q}}
\newcommand{\impactcoordinatechange}{R}

\newcommand{\zerolbl}{\rm zero}
\newcommand{\zdmanifold}{\mathcal{Z}}
\newcommand{\fzero}{f_{\zerolbl}}
\newcommand{\deltazero} [1][] {\delta_{{\zerolbl}#1}}
\newcommand{\impactmapzero} [1][] {\impactmap_{{\zerolbl}#1}}
\newcommand{\zdvfieldlbl}{\kappa}
\newcommand{\zdcoord}{\xi}
\newcommand{\zdone} [1][] {\zdvfieldlbl_{1#1}}
\newcommand{\zdtwo} [1][] {\zdvfieldlbl_{2#1}}
\newcommand{\KEzero}{\KE_{\zerolbl}}
\newcommand{\PEzero}{\PE_{\zerolbl}}
\newcommand{\PEzeromax}{\PEzero^{\max{}}}

\newcommand{\poincaresection}{\mathcal{S}}

\newcommand{\parama}{\alpha}

\newcommand{\leftarray}{\left[\begin{array}}
\newcommand{\rightarray}{\end{array}\right]}


%% file: sections/1-Introduction.tex
Models of bipedal robots\index{robot!bipedal} are hybrid, nonlinear, and typically, high dimensional. In
addition, as it will be motivated shortly, the continuous portion of the
dynamics is often underactuated. A further complication is
that a steady walking cycle is a non-trivial periodic
motion\index{periodic orbit}. This means that standard stability
tools for static equilibria do not apply. Instead, one must use
tools appropriate for the study of periodic orbits, such as
\Poincare\ return maps. The overall complexity in the modelling and analysis of bipedal locomotion has in turn motivated a host of gait design methods that are built around low-dimensional approximations to the dynamics, often based on approximating the system as an inverted pendulum, and also often approximating the legs as massless.


This article outlines an approach to gait design that applies to high-dimensional underactuated models, without making approximations to the dynamics. As with approximate design methods, lower dimensional models do appear, but unlike approximate design methods, the lower dimensional models are exact, meaning that solutions of the low-dimensional model are also solutions of the high-dimensional model evolving in a low-dimensional invariant surface. To get an initial sense of what this may mean, consider a floating-base model of a bipedal robot, and then consider the model with a point or link of the robot, such as a leg end or foot, constrained to maintain a constant position respect to the ground. The given contact constraint is holonomic and constant rank, and thus using Lagrange multipliers (from the principle of virtual work), a reduced-order model compatible with the (holonomic) contact constraint is easily computed. When computing the reduced-order model, no approximations are involved, and solutions of the reduced-order model are solutions of the original floating-base model, with inputs (ground reaction forces and moments) determined by the Lagrange multiplier.

 Virtual constraints are relations (i.e., constraints) on the state variables of the robot's model that are achieved through the action of actuators and feedback control instead of physical contact forces. They are called \textit{virtual} because they can be re-programmed on the fly without modifying any physical connections among the links of the robot or its environment. Virtual constraints can be used to synchronize the evolution of a robot's links to create stable periodic motion. Like physical constraints, under certain regularity conditions, they induce a low-dimensional invariant model, called the \textit{zero dynamics}, due to the highly influential paper \cite{BYRNESC91}. Each virtual constraint imposes a relation between joint variables, and by differentiation a relation between joint velocities. As a consequence, for the cases studied in this chapter, the dimension of the zero dynamics is the initial number of states in the robot's model minus twice the number of virtual constraints (which can be at most the number of independent actuators). The main novelty required for the study of this reduced-dimensional constrained system in bipedal locomotion arises from the \textit{hybrid} or multi-phase nature of locomotion models, such as alternating single support phases and impacts. The hybrid nature of the models gives rise to the \textit{hybrid zero dynamics} (HZD).

 Virtual constraints and hybrid zero dynamics originated in the study of underactuated, planar bipedal locomotion in \cite{GRABPL01,WEGRKO03}; a synthesis of these methods can be found in \cite{WGCCM07,sadati2012hybrid}, with many experiments and extensions reported in  \cite{CHetal03,WEBUGR04,ChDjGr08,yang2009framework,SrPaPoGr2011,PaRaGr13,SrPaPoGr14,martin2014effects,zhao2014human,ames2014human} . The methods are currently being developed for and applied to underactuated 3D robots; see \cite{CHGRSHIH09,WaChTl13,RAHUAKGR14,GrChAmSi2014,AKBUGR14,AKGR14,Zutven:PHD} with experiments just beginning to be reported \cite{BURAHAGRGAGR14,GrizWebYouTube2015}. Virtual constraints and hybrid zero dynamics are also being used in the control of lower-limb prostheses \cite{gregg2014evidence,gregg2014virtual,HoReZhPaAm2015}.

%% file: sections/2-Underactuation.tex
An important source of complexity in a bipedal robot is the degree of
actuation of the model with respect to the number of degrees of freedom. When there are fewer independent actuators than degrees of freedom, the model is \emph{underactuated}\index{underactuation}. It is a common theme in robotics that underactuated models are more challenging to control than fully actuated models.

The primary motivation of this chapter is the \textit{effective  underactuation} that arises in a humanoid robot. Due to the finite size of its feet and the unilateral nature of the contact forces between the foot and ground, most humanoids can all too easily roll forward on the foot, creating an axis of rotation without actuation. Because feet are typically narrower than they are long, it is even easier to roll laterally on the foot. Keeping the foot flat on the ground is difficult and imposes severe restrictions on ankle torque, which is what is meant by ``effective underactuation''. This chapter studies the case when the size of the feet are to a limiting value of zero, namely point feet, resulting in a truly underactuated model. Control designs that allow a robot with point feet to achieve asymptotically stable walking gaits are developed.

Focusing on underactuation is important for at
least two reasons. On one hand, it is interesting to prove, both theoretically and
experimentally, that elegant walking and running motions are
possible with a mechanically simple robot (no feet). On the other
hand, if human walking is taken as the defacto standard against
which mechanical bipedal walking is to be compared, then the
flat-footed walking achieved by current robots needs to be improved.
In particular, toe roll toward the end of the single support phase
needs to be allowed as part of the gait design. Currently, this is
 specifically avoided because, as mentioned above, it leads to
underactuation, namely, \textit{rotation of the foot about the toe introduces an axis of rotation with no actuation, as does lateral rotation of the foot}. A nominal gait design method that produces feasible stable motions without \textit{requiring the use of ankle torque}, can always produce stable motions when powered ankles are available, even in cases where the torque that can be produced may be very limited.  Underactuation and ``effective underactuation'' (severe bounds on ankle torque) are extremely challenging for a control design philosophy based on trajectory
tracking and a quasi-static stability criterion, such as the Zero
Moment Point (ZMP) \cite{VUBOSUST90}, as is currently practiced
widely in the bipedal robotics community. Moreover, the results developed in this article for point feet can easily be extended to the case of finite size foot \cite{ChDjGr08,WaChTl13}. Actuation at the foot can be used to improve the convergence toward a periodic motion or to achieve a human-like evolution of the ZMP.

There is considerable freedom when choosing an underactuated model. One passive joint can be considered in the sagittal plane (at ankle or toe) for studying planar \cite{CHetal03} or 3D biped \cite{WaChTl13,Zutven:PHD} gaits. Two passive joints, one in the sagittal plane and another in the frontal plane, can be introduced for 3D bipeds \cite{CHGRSHIH09} when yaw rotation is assumed to be avoided by friction. Three passive joints are introduced to model point feet in \cite{ShGrCh12}. In the present chapter, all of these sources of underactuation are handled. 

%% file: sections/3-HybridModel.tex
This section introduces a hybrid dynamic model for walking motions of
a bipedal robot. A hybrid model is required because walking consists of alternating phases of one foot on the ground and two feet on the ground.

\begin{figure}[bt]
  \psfrag{test}{\footnotesize $\begin{array}{c} p_2^z(qx) = 0\\[-3pt]\text{and}\\[-3pt] \dot{p}_2^z(q)<0 0\end{array}$}
      \psfrag{impact}{\footnotesize  $x^+ = \impactmap(x^-)$}
      \psfrag{dyn}{\footnotesize  $\dot{x}=f(x) + g(x) u$}
      \centering
      \subfloat[\label{fig:grizzle:modeling:highdof}]{\includegraphics[width=.5\textwidth]{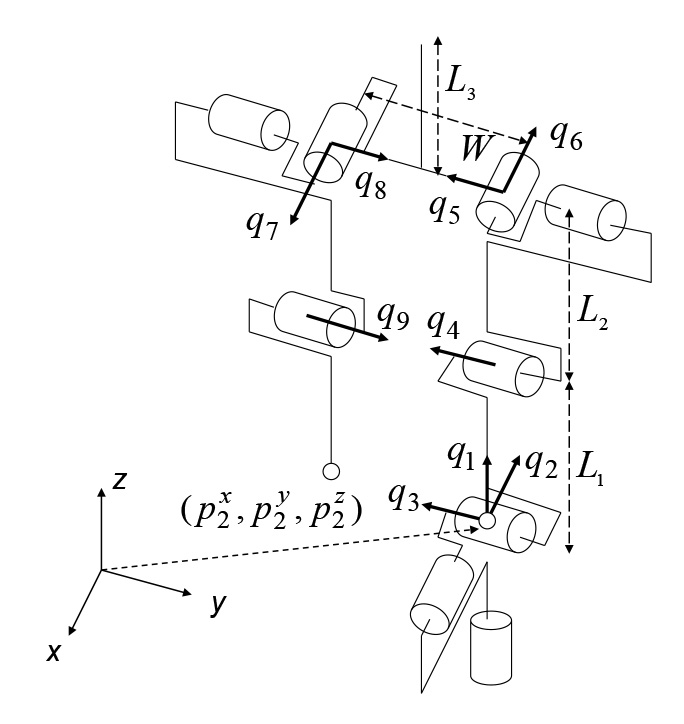}}
      \subfloat[\label{fig:grizzle:modeling:hybrid_walking}]{\includegraphics[width=.5\textwidth]{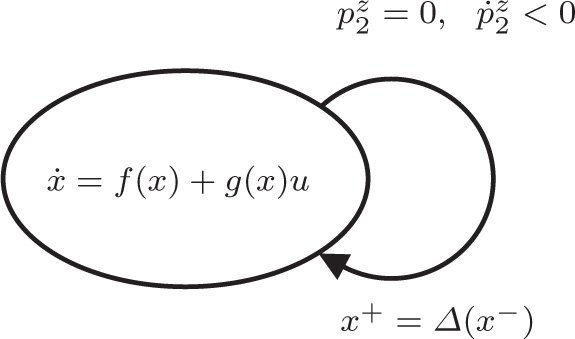}}
    \caption[Typical robot model and a hybrid model of walking.]{(a) A five-link 3D biped
with point feet. The links on the stance leg end have zero length and indicate three passive degrees of freedom: yaw, roll and pitch. Eliminating the bottom link would model a foot with yaw fixed (due to friction, for example), with roll and pitch free. For later use, Cartesian  coordinates $(p_2^x, p_2^y, p_2^z)$ are indicated at the swing leg end. (b) Hybrid model of walking with point feet. Key elements are the continuous dynamics of the single support phase, written in state space form as ${\dot{x}=f(x) + g(x) u}$, the switching or impact condition, $p_2^z(q) = 0$ and $\dot{p}_2^z(q,\dot{q}) < 0$, which detects when the height of the swing leg above the walking surface is zero with the swing leg descending, and the re-initialization rule coming from the impact map, $\impactmap$.}
    \label{fig:grizzle:modeling:highdof_and_hybrid_walking}
  \end{figure}

The robot is assumed to
consist of $\numlinks \ge 2$ rigid links with mass connected via
rigid, frictionless revolute joints. The contact with the ground is modeled as a passive point. A typical robot is depicted in
\fig{grizzle:modeling:highdof}, which is intentionally suggestive of a human
form. All motions are assumed to consist of successive phases of \emph{single support}
(stance leg on the ground and swing leg in the air) and \emph{double
support} (both legs on the ground). The double support phase is assumed to be instantaneous. Conditions that guarantee the
leg ends alternate in ground contact without slipping---while other links such as the
torso or arms remain free---must be imposed during control design. A
rigid impact is used to model the contact of the swing leg with the
ground.

The distinct phases of walking naturally lead to mathematical models
that are comprised of two parts: the differential equations describing
the dynamics during the swing phase and a model that describes the
dynamics when a leg end impacts the ground. For simplicity, in the models developed
here, the ground is assumed to be flat \cite{powell2012motion,GrGr2015ACC}.

\subsection{Lagrangian Swing Phase Model}
\label{sec:grizzle:model:SS}
The swing phase model corresponds to a pinned open kinematic chain.
For simplicity, it is assumed that only symmetric gaits are of interest, and hence it
does not matter which leg end is pinned. The swapping of the roles of
the legs can be accounted for in the impact model.

Let $\Q$ be the $\numlinks$-dimensional configuration manifold of the
robot when the stance leg end is acting as a pivot and let $q:=(q_1;
\cdots; q_\numlinks) \in \Q$ be a set of generalized coordinates. Denote the potential and kinetic
energies by $V(q)$ and $K(q,\dot
q)=\frac{1}{2}\dot{q}^T D(q) \dot q$, respectively, where the inertia
matrix $D$ is positive definite on $\Q$. The dynamic model is easily
obtained with the method of Lagrange\index{Lagrange's method},
yielding the mechanical model
\begin{equation}\label{eqn:grizzle:modeling:mech_model_swing_phase}
  D(q) \ddot{q} + C(q,\dot{q})\dot{q} + G(q) = B u,
\end{equation}
where $u=(u_1; \cdots; u_k) \in \real{k}$ is the vector of input torques, and $B$, the torque distribution matrix, is assumed to be constant with rank $1\le k <N$.  Recall that the foot is modeled as an unactuated point contact, with one, two, or there degrees of freedom.

The model is written in state space form by defining
\begin{eqnarray}
    \dot x & = & \left[
      \begin{array}{c}
        \dot{q}\\
        D^{-1}(q)
        \left[-C(q,\dot{q})\dot{q} - G(q) + B(q) u\right]
      \end{array}\right]\\
  & =: & f(x) + g(x) u \label{eqn:grizzle:modeling:model_walk}
\end{eqnarray}
where $x := (q;\dot{q})$. The state space of the model is $\X =
T\Q$. For each $x\in \X$, $g(x)$ is a $2 \numlinks \times
k$ matrix. In
natural coordinates $(q;\dot{q})$ for $T\Q$, $g$ is independent of
$\dot{q}$.

\subsection{Impact Model}
\label{sec:grizzle:model:DS}
The impact of the swing leg with the ground at the end of a step is
represented with the rigid perfectly inelastic contact model
of \cite{HUMA94}. This model\index{impact model} effectively
collapses the impact phase to an instant in time. The impact forces
are consequently represented by impulses, and a discontinuity or jump
is allowed in the velocity component of the robot's state, with the
configuration variables remaining continuous or constant during the
impact. Since we are assuming a symmetric walking gait, we can avoid
having to use two swing phase models---one for each leg playing the
role of the stance leg---by relabeling the robot's coordinates at
impact. The coordinates must be relabeled because the roles of the
legs must be swapped.  Immediately after swapping, the former swing
leg is in contact with the ground and is poised to take on the role of
the stance leg \cite{WGCCM07,CHGRSHIH09,AKGR14}. The \emph{relabeling} of the generalized coordinates is given by a
matrix, $\impactcoordinatechange$, acting on $q$ with the property
that $\impactcoordinatechange \impactcoordinatechange = I$, i.e.,
$\impactcoordinatechange$ is a circular matrix. The result of the
impact and the relabeling of the states provides an expression
\begin{equation}\label{eqn:grizzle:modeling:delta}
  x^+ = \impactmap(x^-)
\end{equation}
where $x^+:=(q^+; \dot{q}^+)$ (resp. $x^-:=(q^-; \dot{q}^-)$) is the
state value just after (resp. just before) impact and
\begin{equation}\label{eqn:grizzle:modeling:deltadefined}
  \impactmap(x^-) :=
  \left[
    \begin{array}{c}\smallskip
      \impactmappos\, q^- \\ \impactmapvel(q^-)\, \dot{q}^-
    \end{array}
  \right].
\end{equation}
It is noted that the impact map is linear in the generalized velocities. A detailed derivation of the impact map is given in \cite{WGCCM07}.

\subsection{Overall Hybrid Model}
\label{sec:grizzle:model:hybrid}
A hybrid model\index{hybrid model!bipedal walking} of walking is
obtained by combining the swing phase model and the impact model to
form a system with impulse effects\index{system with impulse
effects}. The model is then
%
\begin{equation}\label{eqn:grizzle:modeling:full_hybrid_model_walking}
\Sigma: \begin{cases}
  \begin{aligned}
    \dot x & = f(x) + g(x) u,   & x^- &\notin \poincaresection\\
       x^+ & = \impactmap(x^-), & x^- &\in \poincaresection,
  \end{aligned}
  \end{cases}
\end{equation}
where the switching set is chosen to be
\begin{equation}
  \label{eqn:grizzle:modeling:S}
  \poincaresection := \{ (q,\dot{q}) \in T\Q~|~\pswingfootv(q) = 0,\; \vswingfootv(q, \dot{q}) <0 \}.
\end{equation}
In words, a trajectory of the hybrid model is specified by the swing
phase model until an impact occurs. An impact occurs when the state
``attains'' the set $ \poincaresection$, which represents the walking
surface. At this point, the impact of the swing leg with the walking
surface results in a very rapid change in the velocity components of
the state vector. The impulse model of the impact compresses the
impact event into an instantaneous moment in time, resulting in a
discontinuity in the velocities. The ultimate result of the impact
model is a new initial condition from which the swing phase model
evolves until the next impact.
\Fig{grizzle:modeling:hybrid_walking} gives a graphical representation
of this discrete-event system.

A \emph{step}\index{step} of the robot is a solution of
\eqn{grizzle:modeling:full_hybrid_model_walking} that starts with the robot
in double support, ends in double support with the configurations of
the legs swapped, and contains only one impact event.
\emph{Walking}\index{walking} is a sequence of steps.

%% file: sections/4-PhysicalConstraintsJWG.tex
This section reviews how holonomic constraints in a standard Lagrangian model without control inputs induce a reduced-order (exact) model. Consider a Lagrangian system
\begin{equation}
\label{eqn:grizzle:LagrangeNoControl}
D(q) \ddot{q} + C(q,\dot{q})\dot{q} + G(q) = 0,
\end{equation}
with configuration variables $q\in \Q$, an open subset of $\mathbb{R}^N$. A \textit{regular holonomic constraint} is a twice continuously differentiable  function $h:\Q \to \mathbb{R}^k$
such that
\begin{equation}
\label{eqn:grizzle:ConstraintManifold}
\tilde{\Q}=\left\{q \in \Q~|~h(q) = 0  \right\}
\end{equation}
is non-empty and for each $q_0\in \tilde{\Q}$
\begin{equation}
\label{eqn:grizzle:ConstantRank}
\mathrm{rank}~\frac{\partial h (q_0)}{\partial q}=k.
\end{equation}

The model \eqref{eqn:grizzle:LagrangeNoControl} has a natural restriction to $T\tilde{\Q}$ as a mechanical system with $(N-k)$ DOF. To compute it, the principle of virtual work says to augment \eqref{eqn:grizzle:LagrangeNoControl} with a Lagrange multiplier \begin{equation}
\label{eqn:grizzle:LagrangeNoControlWithMultiplier}
D(q) \ddot{q} + C(q,\dot{q})\dot{q} + G(q) = \left(\frac{\partial h(q)}{\partial q}\right)^T \lambda,
\end{equation}
where $\lambda$ is obtained by setting the second derivative of the constraint along solutions of the model to zero, that is,
\begin{align}
0 &= \frac{d^2h}{dt^2} \nonumber \\
\label{eqn:grizzle:SetAccelerationZero}
&= \frac{\partial h(q)}{\partial q} \ddot{q} + \frac{\partial }{\partial q}\left( \frac{\partial h(q)}{\partial q}\dot{q} \right)\dot{q}.
\end{align}
Indeed, substituting the model \eqref {eqn:grizzle:LagrangeNoControlWithMultiplier} into \eqref{eqn:grizzle:SetAccelerationZero} and solving for the Lagrange multiplier yields
\begin{equation}
\label{eqn:grizzle:LagrangeMultiplier}
 \lambda^\ast =\Lambda^{-1} (q)\left[ \frac{\partial h(q)}{\partial q}D^{-1}(q) \Big( C(q,\dot{q})\dot{q} + G(q) \Big) - \frac{\partial }{\partial q}\left( \frac{\partial h(q)}{\partial q}\dot{q} \right)\dot{q} \right],
\end{equation}
where
\begin{equation}
\label{eqn:grizzle:LagrangeMultiplierSecondPart}
\Lambda(q)=\left(\frac{\partial h(q)}{\partial q} D^{-1}(q) \frac{\partial h(q)}{\partial q}^T \right)
\end{equation}
is \textit{automatically invertible} from the regularity of $h$ and the positive definiteness of $D(q)$.

 The expression for the reduced-order model can be made more explicit if we assume for simplicity that
 the holonomic constraint can be expressed as
\begin{equation}
\label{eqn:grizzle:HolonomicConstraintInq1q2}
0 = \qconstrained - h_d(\qfree),
\end{equation}
 for a choice of configuration variables $(\qconstrained,\qfree)$, with the constrained coordinates $\qconstrained \in \Q_{\rm c} \subset \mathbb{R}^k$, the free coordinates $\qfree \in \Q_{\rm f} \subset \mathbb{R}^{N-k}$, and such that a diffeomorphism $F:\Q_{\rm c} \times  \Q_{\rm f} \to \Q$ exists.
  In this case, the constraint is always regular and the mapping $F_c:  \Q_{\rm f} \to \Q$ given by
 \begin{equation}
F_c(\qfree):= F(h_d(\qfree),\qfree)
\end{equation}
is an embedding; moreover, its image defines the constraint manifold in the configuration space, namely
\begin{equation}
\label{eqn:grizzle:ConstraintManifoldInq1q2}
\tilde{\Q}=\left\{q \in \Q~|~ q = F_c(\qfree),~\qfree \in Q_{\rm f}  \right\},
\end{equation}
which is diffeomorphic to $\Q_{\rm f}$ under $F_c$.

Applying the configuration constraint along a trajectory of the model yields a constraint on velocity, which, using the chain rule, can be written as
\begin{align}
\label{eqn:cc:ConstraintVelocity}
\dot{q} & = \frac{\partial F_c(\qfree)}{\partial \qfree} \dqfree\\
&=: J_c(\qfree) \dqfree.
\end{align}
In addition, using the chain rule again, there is a constraint on acceleration,
\begin{align}
\label{eqn:cc:ConstraintAcceleration}
\ddot{q}&=J_c(\qfree) \ddqfree+ H_c(\qfree,\dqfree),
\end{align}
where $H_c(\qfree,\dqfree)$ contains quadratic terms in velocity resulting from the derivative.

Taking into account the constraints on configuration, velocity and acceleration, the reduced-order model on $T\Q_{\rm f}$ can be written as
\begin{equation}
\label{eqn:grizzle:ReducedMechanicalModel}
\tilde{D}(\qfree) \ddqfree + \tilde{H}(\qfree, \dqfree)=0,
\end{equation}
where
\begin{align}
 \tilde{D}(\qfree) &:= \left. J_c(\qfree)^T D(q) J_c(\qfree) \right|_{ q=F_c(\qfree)}, \nonumber \\
 \tilde{H}(\qfree, \dqfree) &:= \left. J_c(\qfree)^T  \Big(H(q, \dot q) +  D(q) H_c(\qfree,\dqfree) \Big) \right|_{\begin{array}{l} q=F_c(\qfree)\\\dot{q}=J_c(\qfree) \dqfree\end{array}}, \nonumber
 \end{align}
and
$$H(q, \dot q):=C(q,\dot q) \dot q + G(q).$$
The right-hand side of the dynamic model \eqref{eqn:grizzle:ReducedMechanicalModel} is zero because the constraints being satisfied,
$$0=h\circ F_c(\qfree),$$
implies that
\begin{equation}
\label{eqn:grizzle:OrthogonalConditionOnConstraint}
0=\left. \frac{\partial h(q)}{\partial q} J_c(\qfree)\right|_{q=F_c(\qfree)}.
\end{equation}

The solutions of the reduced-order model \eqref{eqn:grizzle:ReducedMechanicalModel} are (exact) solutions to the full-order model \eqref{eqn:grizzle:LagrangeNoControlWithMultiplier}. Indeed, if $(\qfree(t);\dqfree(t))$ is a solution of \eqref{eqn:grizzle:ReducedMechanicalModel}, then
\begin{align}
q(t)  &= F_c(\qfree(t)) \nonumber \\
\dot{q}(t)&= J_c(\qfree(t)) \dqfree(t) ) \nonumber
\end{align}
is a solution of \eqref{eqn:grizzle:LagrangeNoControlWithMultiplier} with the Lagrange multiplier equal to
the unique solution of \eqref{eqn:grizzle:SetAccelerationZero}.

%% file: sections/5-VirtualConstraintsAndZeroDynamics_v02.tex
Physical constraints in a mechanism guide the motion along a constraint surface, but do not impose a specific evolution with respect to time. This property seems particularly well adapted to locomotion. First of all, any attempt to describe walking, even something as simple as the
difference between human-like walking (knees bent forward) and
bird-like walking (knees bent backward), inevitably leads to a
description of the \emph{posture} or \emph{shape} of the robot
throughout a step. In other words, a description of walking involves
at least a partial specification of the path followed in the
configuration space of the biped.

Secondly, in a controller based upon tracking of a time trajectory, if a disturbance were to affect the robot and
causes its motion to be retarded with respect to the planned motion, the feedback system
is then obliged to play catch up in order to regain synchrony with the reference trajectory. Presumably,
what is more important is the orbit of the robot's motion, that is, the path in state space traced out
by the robot, and not the slavish notion of time imposed by a reference trajectory (think about how
you respond to a heavy gust of wind when walking). A preferable situation, therefore, would be for
the control system in response to a disturbance to drive the motion back to the periodic orbit, but not to attempt
otherwise re-synchronizing the motion with respect to time.

With these motivations in mind, a controller based on virtual constraints is now presented. This section parallels the developments in the previous section, but this time for a model with actuation. In particular, the action of contact forces and moments is replaced by actuator torques or forces. The imposed constraints will be virtual because they exist as lines of code in an embedded controller and can be modified on the fly without any physical changes to the robot. While the constraints imposed are virtual, in the authors' opinion, they are as natural as physical constraints.

\subsection{Virtual constraints}
\label{sec:grizzle:ZD:VC}

Since the robot has $k$ independent actuators, $k$ virtual constraints can be generated by the actuators. The virtual constraints are expressed as outputs applied to the model \eqref{eqn:grizzle:modeling:model_walk}, and a feedback controller must be designed that drives asymptotically to zero the output function. In order to emphasize the parallels between virtual and physical constraints, the output is written as
\begin{equation}\label{eqn:grizzle:FdbkDesignApproach:output_fcn_structure_new_coord}
  y = h({q}) := \aq-h_d(\uq),
\end{equation}
where $\aq \in \Q_{\rm c} \subset \mathbb{R}^k$, and $\uq \in \Q_{\ulbl} \subset \mathbb{R}^{N-k}$ with $(\aq, \uq)$ forming  a set of generalized configuration variables for the robot (i.e., such that a diffeomorphism $F:\Q_{\rm c} \times  \Q_{\ulbl} \to \Q$ exists). The interpretation is that $\aq$ represents a collection of variables that one wishes to ``control'' or ``regulate'', while $\uq$ is a complementary set of variables that remain ``free''. Later, a special case of $h_d(\uq)$ will be introduced that highlights a \textit{gait phasing variable}, which makes it easier to interpret the virtual constraint in many instances.

The configuration constraint surface for the virtual constraints is identical to the case of physical constraints in \eqref{eqn:grizzle:ConstraintManifoldInq1q2}. Adding the velocities to the configuration variables gives the \textit{zero dynamics manifold}
\begin{align}
\label{eqn:grizzle:VC:ZeroDynManifold}
\zdmanifold:&=\{(q,\dot{q}) \in T\Q~| y = h(q) = 0, ~\dot{y}=\frac{\partial h(q)}{\partial q}\dot{q}=0\} \nonumber \\
&=\{(q,\dot{q}) \in T\Q~| q=F_c(\uq), \dot{q}=J_c(\uq) \duq, ~~(\uq, \duq) \in T\Q_{\ulbl} \},
\end{align}
which is diffeomorphic to $T\Q_{\ulbl}$. The terminology ``zero dynamics manifold'' comes from \cite{BYRNESC91}. It is the state space for the internals dynamics compatible with the outputs being identically zero.  For the underactuated systems studied here, the dimension of the zero dynamics manifold is $2 (N-k)$.

The torque  $u^\ast $ required to remain on the zero dynamics manifold is computed by substituting the controlled model \eqref{eqn:grizzle:modeling:mech_model_swing_phase} into \eqref{eqn:grizzle:SetAccelerationZero} and solving for the input yielding $\ddot{y}=0$. This gives
\begin{equation}
\label{eqn:grizzle:VC:eq_torque_star} u^\ast =\Big(\frac{\partial h(q)}{\partial
q}D^{-1}(q)B\Big)^{-1}\left( { - \frac{\partial }{\partial q}\left( \frac{\partial h(q)}{\partial q}\dot{q} \right)\dot{q}+\frac{\partial h(q)}{\partial
q}D^{-1}(q)H(q,\dot {q}) } \right).
\end{equation}
For $u^\ast$ to be well defined, the decoupling matrix
\begin{equation}
\label{eqn:grizzle:VC:decouplingMatrix} A(q): = \frac{\partial h(q)}{\partial
q}D^{-1}(q)B
\end{equation}
must be invertible\footnote{In the control literature, the output is said to have vector relative degree two. More general constraints can be handled.}. In the case of physical constraints, the invertibility of  \eqref{eqn:grizzle:LagrangeMultiplierSecondPart} was automatic as long as the constraint was regular. With virtual constraints, \textit{the invertibility of the decoupling matrix is not automatic} and must be checked. Numerous examples exist, nevertheless, that show it is straightforward to meet this condition. Later in the chapter it is shown how to compute the decoupling matrix without inverting the inertia matrix.

Another difference with physical constraints is that it is possible for the system to be either initialized off the virtual constraint surface or perturbed off it.  Feedback is therefore required to asymptotically drive the state of the robot to the constraint surface. The feedforward term $u^\ast$  can be modified to an input-output linearizing controller  \cite[Chap.~5]{WGCCM07},
\begin{equation}
\label{eqn:grizzle:VC:eq_torque} u=u^\ast - A(q)^{-1}\left(\frac{K_p }{\varepsilon ^2}y+\frac{K_d }{\varepsilon
}\dot {y}\right),
\end{equation}
which results in
\begin{equation}
\label{eqn:grizzle:VC:eq_closed_loop}
    \ddot {y}+\frac{K_d }{\varepsilon }\dot
{y}+\frac{K_p }{\varepsilon ^2}y=0.
\end{equation}
A richer set of feedback controllers based on control Lyapunov functions is given in \cite{ames2014rapidly}.

Like physical constraints, under certain regularity conditions, virtual constraints induce a low-dimensional invariant model of the swing-phase dynamics. The low-dimensional model is called the \textit{zero dynamics} in the nonlinear control literature \cite{BYRNESC91}. The main novelty required in biped locomotion arises from the hybrid nature of the models, which gives rise to the \textit{hybrid zero dynamics} \cite{WEGRKO03}.

The zero dynamics of the hybrid model
\eqn{grizzle:modeling:full_hybrid_model_walking} with output
\eqn{grizzle:FdbkDesignApproach:output_fcn_structure_new_coord} are
developed in a two-step process. First, the zero dynamics of the
(non-hybrid) nonlinear model consisting of the swing phase dynamics
\eqn{grizzle:modeling:model_walk} and the output
\eqn{grizzle:FdbkDesignApproach:output_fcn_structure_new_coord} are
characterized, and then, second, an impact invariance condition is
imposed on the swing-phase zero dynamics manifold to obtain
the hybrid zero dynamics.

\subsection{The Swing Phase Zero Dynamics}
\label{sec:grizzle:ZD:SS}
The objective is to characterize the swing-phase model
\eqref{eqn:grizzle:modeling:mech_model_swing_phase} restricted to the constraint surface \eqref{eqn:grizzle:VC:ZeroDynManifold}. The zero dynamics, by definition, reflects the internal dynamics when the output is identically zero, meaning the system evolves on the constraint surface. The development is analogous to the case of physical constraints, once the invertibility of a key matrix is established.

\begin{proposition} Let $B^\perp$ be a $(N-k) \times N$ matrix of rank $N-k$ such that $B^\perp B=0$ and suppose that the decoupling matrix \eqref{eqn:grizzle:VC:decouplingMatrix} is invertible at a point $q$. Then the following matrices are each $N \times N$ and have rank $N$:
$$\left[ \begin{array}{c} \frac{\partial h(q)}{\partial q}  \\ B^\perp D(q)\end{array} \right], ~
\left[ \begin{array}{c} \frac{\partial h(q)}{\partial q} D^{-1}(q) \\ B^\perp \end{array} \right],~and~~
\left[ \begin{array}{cc} \frac{\partial h(q)}{\partial q} D^{-1}(q) B & ~~~~\frac{\partial h(q)}{\partial q} D^{-1}(q) \left(B^\perp \right)^T\\ 0 & B^\perp  \left(B^\perp \right)^T \end{array} \right],~
$$
\end{proposition}

The proof is sketched. Multiplying the leftmost matrix by the inverse of $D(q)$ gives the matrix in the middle. Multiplying the matrix in the middle by the full rank matrix $[B~|\left(B^\perp \right)^T]$ gives the rightmost matrix. This matrix is full rank because it is block upper triangular with the upper left block being the decoupling matrix, which has full rank by assumption, and the lower right block has full rank by standard properties of matrices. This completes the proof.

Multiplying the leftmost matrix by the Jacobian of $F_c:  \Q_{\ulbl} \to \Q$ and restricting to the constraint surface gives
$$\left. \left[ \begin{array}{c} \frac{\partial h(q)}{\partial q} J_c(\uq) \\ B^\perp D(q) J_c(\uq)\end{array} \right] \right|_{ q=F_c(\uq)} = \left. \left[ \begin{array}{c} 0  \\ B^\perp D(q)J_c(\uq)\end{array} \right]  \right|_{ q=F_c(\uq)},
$$
where \eqref{eqn:grizzle:OrthogonalConditionOnConstraint} has been used. The proposition and the fact that $J_c(\uq)$ has rank $N-k$ give the following result.

\begin{corollary}
\label{cor:grizzle:InverseExistence}
Let $B^\perp$ be a $(N-k) \times N$ matrix of rank $N-k$ such that $B^\perp B=0$ and suppose that the decoupling matrix \eqref{eqn:grizzle:VC:decouplingMatrix} is invertible at a point $q=F_c(\uq)$. Then the matrix
\begin{equation}
\label{eqn:grizzle:VC:BperpDjacMatrix}
\left. M(\uq):=B^\perp D(q) J_c(\uq) \right|_{ q=F_c(\uq)},
\end{equation}
is invertible.
\end{corollary}

Armed with the corollary, the calculation of the reduced-order model associated with the virtual constraints is very similar to the case of physical constraints. Multiplying the controlled system \eqref{eqn:grizzle:modeling:mech_model_swing_phase} on the left by $B^\perp$ as in \cite{shiriaev2008can,shiriaev2010transverse} gives
\begin{equation}
\label{eqn:CC:dym_zeroStep1}
  B^\perp D(q) \ddot{q} + B^\perp H(q,\dot{q}) = 0.
\end{equation}
Using \eqref{eqn:cc:ConstraintAcceleration} and restricting to $\zdmanifold$ gives
\begin{equation}\label{eqn:CC:dym_zeroStep2}
  \left.B^\perp D(q) J_c(\uq) \ddot{\uq} + B^\perp \left[  D(q) H_c(\uq,\dot{\uq}) + H(q,\dot{q}) \right]= 0 \right|_{\begin{array}{l} q=F_c(\uq)\\ \dot{q}=J_c(\uq) \dot{\uq}\end{array}},
\end{equation}
yielding a second-order model analogous to \eqref{eqn:grizzle:ReducedMechanicalModel},
\begin{equation}
\label{eqn:CC:dym_zeroStep3}
  M(\uq) \ddot{\uq} + H_{zero}(\uq,\dot{\uq}) = 0,
\end{equation}
where
$$ H_{zero}(\uq,\dot{\uq}):= \left. B^\perp \left[  D(q) H_c(\uq,\dot{\uq}) + H(q,\dot{q}) \right] \right|_{\begin{array}{l} q=F_c(\uq)\\ \dot{q}=J_c(\uq) \dot{\uq}\end{array}}.$$

From this equation, a state variable model for the zero dynamics can be obtained. Selecting $$z=\left[\begin{array}{c} z_1\\ z_2 \end{array}\right]:=\left[\begin{array}{c} \uq\\ \duq \end{array}\right]$$
gives
\begin{align}
  \dot z & =    \left[
      \begin{array}{c}
        z_2\\
        - M^{-1}(z_1) H_{zero}(z_1,z_2)
      \end{array}\right]\\
  & =: f_{zero}(z).  \label{eqn:CC:modeling:model_zero}
\end{align}

An alternative form for the zero dynamics can be obtained from the Lagrangian nature of \eqref{eqn:grizzle:modeling:mech_model_swing_phase}. Let
$$L(q,\dot{q})=\frac{1}{2} \dot{q}^T D(q) \dot{q}- V(q)$$
denote the Lagrangian of the model. Then from Lagrange's equation
$$ \frac{d}{dt} B^\perp \frac{\partial L}{\partial \dot{q} } =  B^\perp  \frac{\partial L}{\partial q },$$
because $B^\perp B u=0$. The term $B^\perp \frac{\partial L}{\partial \dot{q} }$ is a form of \textit{generalized angular momentum}. Restricting to the zero dynamics manifold gives
$$  \dusigma  = \kappa(\uq,\duq),$$
where
\begin{align} \usigma&:=M(\uq) \duq \nonumber \\
\kappa(\uq,\duq) &:= \left. B^\perp  \frac{\partial L(q,\dot{q}) }{\partial q } \right|_{
\begin{array}{l} q=F_c(\uq)\\ \dot{q}=J_c(\uq) \dot{\uq}\end{array}
}. \nonumber
\end{align}
Choosing the zero dynamics coordinates as
$$z=\left[\begin{array}{c} z_1\\ z_2 \end{array}\right]:=\left[\begin{array}{c} \uq\\ \usigma \end{array}\right]$$
gives
\begin{align}
  \dot z & =    \left[
      \begin{array}{c}
        M^{-1}(z_1) z_2\\
        \bar{\kappa}(z_1,z_2)
      \end{array}\right] \label{eqn:CC:modeling:model_zeroAlternative}\\
  & =: f_{zero}(z),   \nonumber
\end{align}
where $\bar{\kappa}(z_1,z_2) = \kappa(z_1, M^{-1}(z_1)z_2)$.\\

\subsection{The Hybrid Zero Dynamics}
\label{sec:grizzle:ZD:Hybrid}

\begin{figure}[t!]
  \centering 
  \subfloat[]{\includegraphics[width=.4\textwidth]{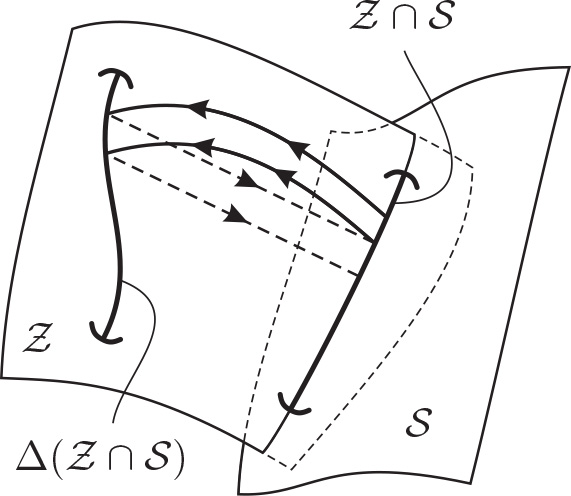}} \hspace{.5cm}
  \subfloat[]{\includegraphics[width=.4\textwidth]{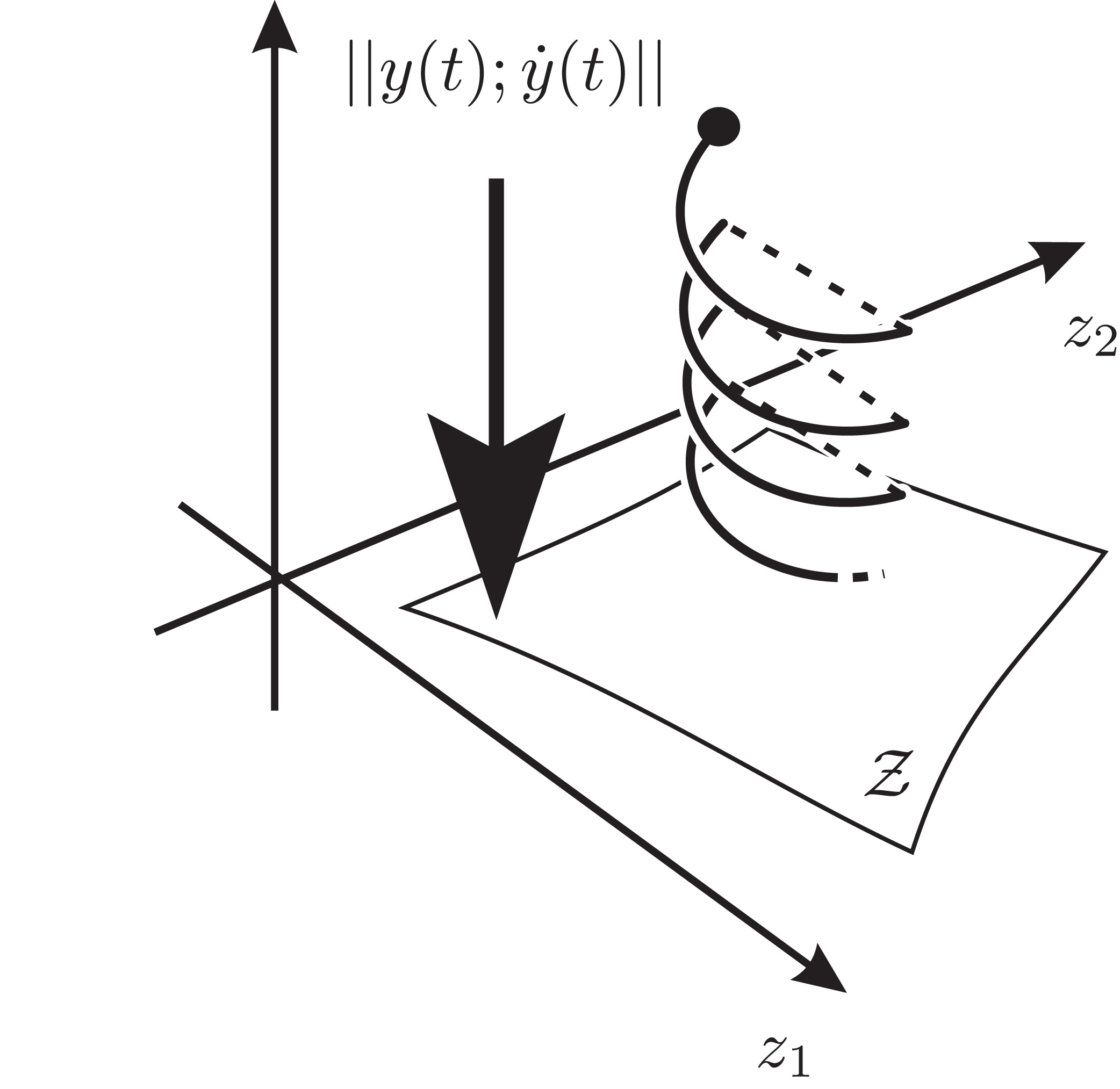}}
  \caption{In (a) hybrid invariant manifold.  In (b), as $||y(t);\dot{y}(t)||$ tends to zero, the solution of the closed-loop system approaches the zero dynamics manifold, $\zdmanifold$.}
  \label{fig:grizzle:ZDand}
\end{figure}

To obtain the hybrid zero dynamics, the zero dynamics manifold must be
\emph{invariant under the impact map}, that is
\begin{equation}\label{eqn:grizzle:impact_invar_cond}
  \impactmap (\poincaresection \cap \zdmanifold) \subset \zdmanifold.
\end{equation}
This condition means that when a solution evolving on $\zdmanifold$ meets the switching surface, $\poincaresection$, the new initial condition arising from the impact map is once again on $\zdmanifold$.

\begin{definition}[Hybrid zero dynamics \cite{WEGRKO03}]
Consider the hybrid model \eqref{eqn:grizzle:modeling:full_hybrid_model_walking} and
output \eqref{eqn:grizzle:FdbkDesignApproach:output_fcn_structure_new_coord}.
Suppose that the decoupling matrix
\eqref{eqn:grizzle:VC:decouplingMatrix} is invertible and
let $\zdmanifold$ and $\dot z = \fzero(z)$ be the associated zero
dynamics manifold and zero dynamics of the swing phase model.  Suppose
that $\poincaresection \cap \zdmanifold$ is a smooth, co-dimension one,
embedded submanifold of $\zdmanifold$. Suppose furthermore that $\impactmap
(\poincaresection \cap \zdmanifold) \subset \zdmanifold.$ Then the
nonlinear system with impulse effects,
\begin{equation}\label{eqn:grizzle:zero_dynamics:reduced_hybrid_model}
  \Sigma_\zero: \begin{cases}
    \begin{aligned}
      \dot z & = \fzero(z), &    z^- & \notin \poincaresection\cap \zdmanifold\\
      z^+ & = \impactmapzero(z^-), & z^- & \in \poincaresection\cap \zdmanifold,
    \end{aligned}
    \end{cases}
\end{equation}
with $\impactmapzero:= \left. \impactmap \right|_{\poincaresection\cap \zdmanifold}$, is the \emph{hybrid zero
dynamics}\index{zero dynamics!hybrid}\index{hybrid zero dynamics}.
\end{definition}

The invariance condition \eqref{eqn:grizzle:impact_invar_cond} is equivalent to
\begin{align}
0=&h\circ \impactmappos(q) \\
0=& \left. \frac{\partial h(\bar{q})}{\partial q} \right|_{\bar{q}=\impactmappos(q)} \impactmapvel(q)\, \dot{q}
\end{align}
for all $(q;\dot{q})$ satisfying
\begin{align}
 h(q) =0, ~~\frac{\partial h(q)}{\partial q} \dot{q} =& 0 \\
\pswingfootv(q) = 0,~~\vswingfootv(q, \dot{q})<& 0.
\end{align}
At first glance, these conditions appear to be very hard to meet. In the case of models with one degree of underactuation (i.e., $k=N-1$), however, it is known that if a single non-trivial solution of the zero dynamics satisfies these conditions, then all solutions of the zero dynamics will satisfy them \cite[Thm.~5.2]{WGCCM07}. In the case of systems with more than one degree of underactuation, systematic methods have been developed which modify the virtual constraints ``at the boundary'' and allow the conditions to be met \cite{MOGR08}. Very straightforward implementations of the result are presented in a robotics context in \cite{CHGRSHIH09} and \cite{GrGr2015ACC}.

\noindent \textbf{Remark:} When low-dimensional pendulum models are used for gait design, they \textit{approximate} the swing phase dynamics and the impact map is ignored. The zero dynamics is an \textit{exact} low-dimensional model that captures the underactuated nature of the robot. Presumably, one could use it without insisting on the impact invariance condition \eqref{eqn:grizzle:impact_invar_cond}. To the knowledge of the authors, this has not been done.

\subsection{Calculating the Feedback Controller}
\label{sec:grizzle:ZD:Practical}

An alternative expression for the torque $u^\ast$ on the zero dynamics manifold can be given. Let $B^+$ be the pseudoinverse of the full rank matrix $B$. Then
\begin{equation}
\label{eqn:grizzle:VC:ustarFromZD}
 u^\ast = B^+ D(q) J_c(\uq) \ddot{\uq} + B^+ \left[H(q,\dot{q}) + D(q) H_c(\uq,\dot{\uq}) \right] \bigg|_{\begin{array}{l} q=F_c(\uq)\\\dot{q}=J_c(\uq) \dot{\uq}\end{array}} .
\end{equation}
In other words, the control signal required to remain on the zero dynamics manifold can be recovered directly from knowledge of the solution to any one of \eqref{eqn:CC:dym_zeroStep3}, \eqref{eqn:CC:modeling:model_zero}, or \eqref{eqn:CC:modeling:model_zeroAlternative}. In particular, it is not necessary to explicitly compute the decoupling matrix.

A similar expression can be developed to calculate the torque ensuring convergence toward the zero dynamics surface, once again without explicitly computing the decoupling matrix. The important advantage over \eqref{eqn:grizzle:VC:eq_torque} is that the matrix to invert has dimension equal to the unactuated degrees of freedom in the model, which is typically much smaller than $N$ or $k$.

Suppose that the control objective is
\begin{equation}
\label{eqn:grizzle:VC:eq_closed_loop_a}
    \ddot {y} = \Gamma(y, \dot{y});
\end{equation}
a special case would be \eqref{eqn:grizzle:VC:eq_closed_loop}. By definition of the virtual constraint, the output and its derivatives are
\begin{align}
  y &=  \aq-h_d(\uq), \label{eqn:grizzle:VC:yformula}\\
  \dot y &= \daq - \frac{\partial h_d(\uq)}{\partial \uq} \duq, \label{eqn:grizzle:VC:dyformula}\\
  \ddot y &= \ddaq - \frac{\partial h_d(\uq)}{\partial \uq} \dduq  - \frac{\partial}{\partial \uq} \left(\frac{\partial h_d(\uq)}{\partial \uq} \duq \right)\duq.
  \label{eqn:grizzle:VC:ddyformula}
\end{align}
Using equation \eqref{eqn:grizzle:VC:eq_closed_loop_a} to define $\ddot {y}$, it follows that the acceleration of the controlled variable in closed-loop satisfies
\begin{equation}
\label{eqn:grizzle:VC:ddqcformula}
\ddaq = \frac{\partial h_d(\uq)}{\partial \uq} \dduq + \frac{\partial}{\partial \uq} \left(\frac{\partial h_d(\uq)}{\partial \uq} \duq \right)\duq+ \Gamma(y, \dot{y}),
\end{equation}
where $y$ and $\dot{y}$ are computed in \eqref{eqn:grizzle:VC:yformula} and \eqref{eqn:grizzle:VC:dyformula}.

Turning now to the dynamic model \eqref{eqn:grizzle:modeling:mech_model_swing_phase},
and using
\begin{align}
  q &=  F(\aq,\uq), \label{eqn:grizzle:VC:qformula}\\
  \dot q &=  \frac{\partial F(\aq,\uq)}{\partial \aq} \daq + \frac{\partial F(\aq,\uq)}{\partial \uq} \duq, \label{eqn:grizzle:VC:dqformula}\\
  \ddot q &= \frac{\partial F(\aq,\uq)}{\partial \aq} \ddaq + \frac{\partial F(\aq,\uq)}{\partial \uq} \dduq + \Psi(\aq,\uq,\daq,\duq)
  \label{eqn:grizzle:VC:ddqformula}
  \end{align}
in combination with \eqref{eqn:grizzle:VC:ddqcformula} results in
\begin{equation}
\label{eqn:grizzle:VC:ModelInNewCoordinates}
 \widebar{D}(\aq,\uq)  J_r(\aq,\uq) \dduq + \Omega(\aq,\uq,\daq,\duq,y,\dot{y}) = B u,
\end{equation}
where
\begin{align}\nonumber
\widebar{D}(\aq,\uq) &=  D(q) \bigg|_{q=F(\aq,\uq)} \bigskip \\
J_r(\aq,\uq) &=
  \frac{\partial F(\aq,\uq)}{\partial \aq} \frac{\partial h_d(\uq)}{\partial \uq}+ \frac{\partial F(\aq,\uq)}{\partial \uq} \bigskip \nonumber \\
    \Omega(\aq,\uq,\daq,\duq, y,\dot{y}) &= \widebar{D}(\aq,\uq) \bigg[  \frac{\partial F(\aq,\uq)}{\partial \aq} \bigg( \frac{\partial}{\partial \uq}
  \left(\frac{\partial h_d(\uq)}{\partial \uq} \duq \right)\duq + \bigskip \nonumber \\
&~~~ \Gamma(y, \dot{y}) \bigg)  + \Psi(\aq,\uq,\daq,\duq) \bigg] + \widebar{H}(\aq,\uq,\daq,\duq) \bigskip \nonumber \\
\Psi(\aq,\uq,\daq,\duq)&=\frac{\partial}{\partial (\aq, \uq) } \left( \frac{\partial F(\aq,\uq)}{\partial (\aq, \uq) } \left[\begin{array}{l}  \daq \\ \duq \end{array} \right] \right) \left[\begin{array}{l}  \daq \\ \duq \end{array} \right] \bigskip  \nonumber \\
\widebar{H}(\aq,\uq,\daq,\duq)&=  C(q,\dot{q})\dot{q} + G(q) \bigg|_{\begin{array}{l} q=F(\aq,\uq)\\ \dot{q}=\frac{\partial F(\aq,\uq)}{\partial \aq} \daq + \frac{\partial F(\aq,\uq)}{\partial \uq} \duq \end{array} } \nonumber
\end{align}
Multiplying \eqref{eqn:grizzle:VC:ModelInNewCoordinates} on the left by the full rank matrix
$$\left[
      \begin{array}{c}
        B^\perp \\
        B^+
      \end{array}\right]
      $$
      gives
\begin{align}
        B^\perp \widebar{D}(\aq,\uq)  J_r(\aq,\uq) \dduq +B^\perp \Omega(\aq,\uq,\daq,\duq,y,\dot{y}) &= 0 \\
        B^+ \widebar{D}(\aq,\uq)  J_r(\aq,\uq) \dduq +B^+ \Omega(\aq,\uq,\daq,\duq,y,\dot{y}) &= u.
\end{align}
It follows that a feedback control law achieving \eqref{eqn:grizzle:VC:eq_closed_loop_a} is
\begin{align}
\label {eqn:grizzle:VC:FeedbackEfficient}
 u&= B^+ \widebar{D}(\aq,\uq)  J_r(\aq,\uq) v +B^+ \Omega(\aq,\uq,\daq,\duq,y,\dot{y}) \\
 v &= -\left(B^\perp \widebar{D}(\aq,\uq)  J_r(\aq,\uq)\right)^{-1} B^\perp \Omega(\aq,\uq,\daq,\duq,y,\dot{y}).
\end{align}
This form of the feedback law only requires the inversion of an $(N-k) \times (N-k)$ matrix, where the size corresponds to the number of unactuated coordinates. Corollary \ref{cor:grizzle:InverseExistence} guarantees that the matrix is invertible near the zero dynamics manifold. A uniqueness result in \cite{BYRNESC91} implies that $u$ in \eqref{eqn:grizzle:VC:FeedbackEfficient} when restricted to $\zdmanifold$ is equal to $u^*$ in \eqref{eqn:grizzle:VC:eq_torque_star} and \eqref{eqn:grizzle:VC:ustarFromZD}.

\noindent \textbf{Remark:} Commonly,  $\aq$ and $\uq$ are a linear function of $q$, which then greatly simplifies many of the above equations.

%% file: sections/6-AnalysisZeroDynamics_v03JWG.tex
This section first presents the relation between the stability of a periodic solution in the full-order model and in the hybrid zero dynamics. The stability of periodic orbits within the hybrid zero dynamics is subsequently analyzed.

\subsection{Closed-loop stability with zero dynamics}

Consider a Lipschitz continuous feedback law
\begin{equation}
\label{eqn:grizzle:paramFeedback}
u = \alpha(x,\epsilon)
\end{equation}
that depends on a tuning parameter $\epsilon>0$ and apply it to the model \eqref{eqn:grizzle:modeling:full_hybrid_model_walking}. Let $P^\epsilon : S\to S$ be the Poincar\'e (return) map\footnote{It is in general a partial map and depends on the tuning parameter through the feedback \eqref{eqn:grizzle:paramFeedback}. It is defined by starting with a point in $\poincaresection$, applying the impact map, using this as an initial condition of the closed-loop ODE, and following the flow until it first crosses $\poincaresection$ at time $T_I$; see \cite[Chap.~4]{WGCCM07} for a careful definition.} of the closed-loop system, defined in the usual way. For a given $\epsilon>0$, $x^*\in S$ is a fixed point if $P^\epsilon(x^*)=x^*$, which is well-known to be equivalent to the existence of a periodic solution of the closed-loop system.

The feedback \eqref{eqn:grizzle:paramFeedback} is \textit{compatible with the zero dynamics} if
\begin{equation}
\label{eqn:grizzle:allowedFeedback}
\left. \alpha(x,\epsilon)\right|_{Z} = \left. u^\ast(x)\right|_{Z}.
\end{equation}
In this case, if $\Delta(x^*)\in Z$, then the periodic solution is independent of the tuning parameter $\epsilon$.
Based on \cite{ames2014rapidly}, the feedback is said to \textit{drive the virtual constraints rapidly exponentially to zero} if there exist constants $\beta>0$ and $\gamma>0$ such that, for each $\epsilon>0$, there exists $\delta>0$, such for all $x_0\in B_{\delta}(\Delta(x^*))$,
\begin{equation}
\label{eqn:grizzle:ExpRapidly}
||y(t,x_0), \dot{y}(t,x_0)|| \le \frac{\beta}{\epsilon} e^{-\frac{\gamma}{\epsilon }t} ||y(0),\dot{y}(0)||.
\end{equation}

\begin{theorem}(\cite{MOGR08}, \cite{ames2014rapidly}) \textbf{Determining Closed-loop Stability from the HZD} \label{thm:grizzle:StabilityFromHZD}
Suppose that the virtual constraints \eqref{eqn:grizzle:FdbkDesignApproach:output_fcn_structure_new_coord} satisfy
\begin{enumerate}
\renewcommand{\labelenumi}{(\alph{enumi})}
\item the decoupling matrix is invertible,
\item the associated zero dynamics manifold $\zdmanifold$ is hybrid invariant, and
\item there exists a point $x^*\in S\cap Z$ giving rise to a periodic orbit.
\end{enumerate}
Assume moreover that the feedback \eqref{eqn:grizzle:paramFeedback}
\begin{enumerate}
\renewcommand{\labelenumi}{(\alph{enumi})}
\item[(d)] is compatible with the zero dynamics in the sense of \eqref{eqn:grizzle:allowedFeedback} and
\item[(e)]  drives the virtual constraints rapidly exponentially to zero.
\end{enumerate}
Then there exists $\bar{\epsilon} > 0$ such
that for $0 < \epsilon < \bar{\epsilon}$, the following are equivalent:
\begin{enumerate}
  \renewcommand{\labelenumi}{\roman{enumi})}
      \item the periodic orbit is locally exponentially stable;
  \item $x^*$ is an exponentially stable fixed point of
  $P^{\epsilon};$ and
    \item $x^*$ is an exponentially stable fixed point of $\rho$,
\end{enumerate}
where the restricted Poincar\'e map
\begin{equation}
\label{eqn:grizzle:RestrictedPoincareMap}
\rho :=
P^{\epsilon}|_{\poincaresection \cap \zdmanifold}
\end{equation}
is the Poincar\'e map of the hybrid zero dynamics, is independent of the feedback, and hence is also independent of $\epsilon$.
{\vspace{-0.05in} \begin{flushright} $\Box$ \end{flushright}}
\end{theorem}

\noindent \textbf{Remark:} It is re-emphasized that periodic orbits of the hybrid zero dynamics are periodic orbits of
the full-dimensional model. Two feedback controllers are
provided in \cite{MORRISB05,WEGRKO03} for exponentially stabilizing
these orbits in the full-dimensional model, \eqref{eqn:grizzle:modeling:model_walk}, and a third family of feedback controllers is presented in \cite{ames2014rapidly}.

\subsection{Special Case of One Degree of Underactuation}
\label{sec:grizzle:ZD:Orbits}

When there is only one degree of underactuation, the restricted Poincar\'e map \text{$\rho :=
P^{\epsilon}|_{\poincaresection \cap \zdmanifold}$} is one-dimensional and can be computed in closed form. While one degree of underactuation is primarily a ``planar'' (2D) robot phenomenon, it has also been used in 3D robots \cite{Zutven:PHD} with both single and (nontrivial) double support phases.

The analysis here is from \cite[Chap.~6]{WGCCM07}, which shows that in the case of one-degree of underactuation, the zero dynamics can be written in a particularly simple form. Let $\theta$ be a configuration variable tied to the world frame, such as the absolute angle of the line connecting the stance hip to the end of the stance leg, and let $\sigma$ be the generalized angular momentum conjugate to $\theta$. Then in the coordinates $(\theta, \sigma)$, \eqref{eqn:CC:modeling:model_zeroAlternative} becomes
\begin{subequations}\label{eqn:grizzle:zero_dynamics:zd_states}
 \begin{eqnarray}
   \dot \theta  & = & \zdone(\theta)\sigma \label{eqn:grizzle:zero_dynamics:zd_state_1} \\
   \dot{\sigma} & = & \zdtwo(\theta) \label{eqn:grizzle:zero_dynamics:zd_state_2},
 \end{eqnarray}
\end{subequations}
where $\zdtwo$ is now \textit{independent of angular momentum}. Assume the virtual constraints have been selected such that the zero dynamics admit a periodic orbit, as in Theorem \ref{thm:grizzle:StabilityFromHZD}. It can be shown that $ \dot \theta $ and $\sigma$ will not change signs. Assume furthermore there exists a single point $\theta^-$ such that the swing foot height decreases to zero, with a strictly negative impact velocity in the vertical direction. In this case, \cite[Chap.~6]{WGCCM07} shows that the impact surface in the zero dynamics can be written
\begin{equation}
  \poincaresection \cap \zdmanifold  =
  \left\{(\theta^-;\sigma^-)~|~\sigma^- \in \real{}\right\}.
\end{equation}
For $(\theta^-;\sigma^-) \in \poincaresection \cap \zdmanifold$, let
$$(\theta^+;\sigma^+) = \impactmapzero(\theta^-;\sigma^-).$$

The impact map of Sect. \ref{sec:grizzle:model:DS}, which is based on \cite{HUMA94}, respects conservation of angular momentum (see \cite[Eqn.~(3.20)]{WGCCM07}). It follows that
\begin{equation}
\label{grizzle:zero_dynamics:AngularMomentumTransfer0}
\sigma^+=\sigma^- + m d \dot z^-
\end{equation}
where $d$ is the distance between the feet (measured along the $x$-axis) and $\dot{z}$ the vertical velocity of the center of mass just before impact. In general, $\dot{z}$ is linear with respect to joint velocities, and on $\zdmanifold$, it is therefore proportional to $\dot{\theta}$. Hence,  using \eqref{eqn:grizzle:zero_dynamics:zd_state_1} and the chain rule,
\begin{align}
\label{grizzle:zero_dynamics:AngularMomentumTransfer1}
\sigma^+&=\sigma^- + m d \frac{d z(\theta^-)}{d \theta} \zdone(\theta^-)\sigma^- \\
\label{grizzle:zero_dynamics:AngularMomentumTransfer2}
\sigma^+&=: \deltazero \sigma^-,
\end{align}
where it is noted that $\deltazero<1$ when $\dot z^-<0$, that is, when the vertical velocity of the center of mass is directed downward at the end of the step.

\begin{figure}[bt]
    \centerline{%
    \includegraphics[scale=1.5]{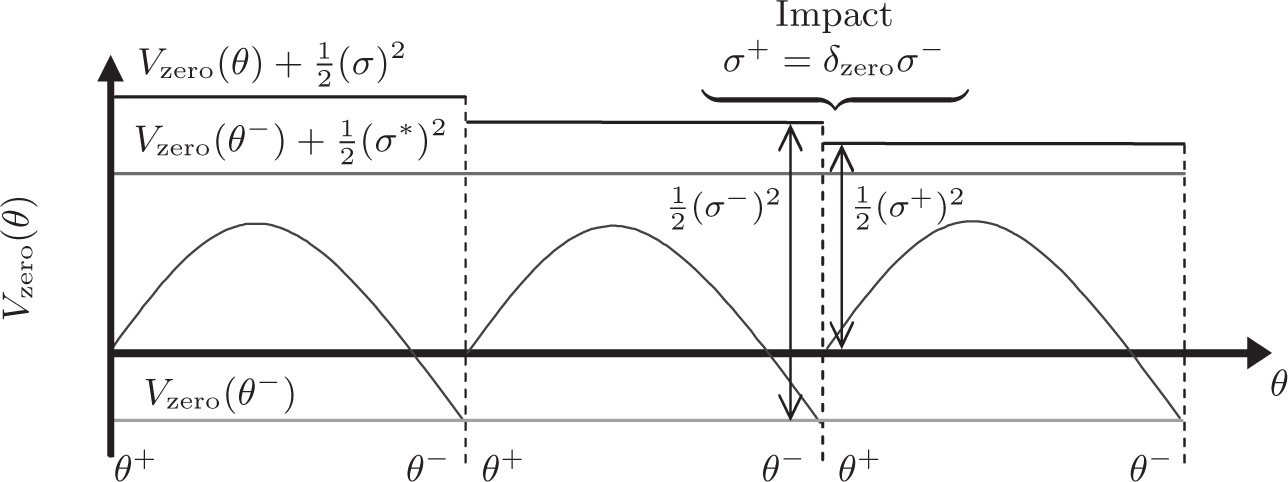}}
    \caption
    {A qualitative look at stability through pseudo energy \cite{WGCCM07}. The zero dynamics is Lagrangian, and thus throughout the single support phase, the corresponding total energy $\PEzero(\theta)+ \frac{1}{2} {\sigma}^2$ is constant. At impact, the change in total energy depends on the angular momentum through $\deltazero {\sigma}^-$ and the potential energy through $\PEzero(\theta^-)$. The total energy corresponding to the periodic orbit is  $\PEzero(\theta^-)+ \frac{1}{2} ({\sigma}^*)^{2}$. Convergence to this total energy level occurs if the angular momentum decreases during impact, namely, $\deltazero < 1$. From the expression for the existence of a periodic orbit, $\deltazero<1 $ is equivalent to $\PEzero(\theta^-) < 0$.}
    \label{fig:within_stride_ctrl:walking_energytransfer}
  \end{figure}

%
The hybrid zero dynamics is thus given by
\eqn{grizzle:zero_dynamics:zd_states} during the swing phase, and at impact
with $\poincaresection \cap \zdmanifold$, the re-initialization rule
\eqref{grizzle:zero_dynamics:AngularMomentumTransfer2} is applied. For $\theta^+ \le \zdcoord_1 \le \theta^- $, define
\begin{equation}\label{eqn:grizzle:zero_dynamics:Vzero}
  \PEzero(\theta) :=
  -\int_{\theta^+}^{\theta}\frac{\zdtwo(\zdcoord)}{\zdone(\zdcoord)}\,
  d\zdcoord.
\end{equation}
A straightforward computation shows that ${\cal L}_{\rm zero} :=
\KEzero - \PEzero$ \cite{WEGRKO03}, where
\begin{equation}\label{eqn:grizzle:zero_dynamics:lagrangian}
  \KEzero =
  \frac{1}{2} \sigma ^2,
\end{equation}
is a Lagrangian\index{zero dynamics!Lagrangian} of the swing-phase
zero dynamics \eqn{grizzle:zero_dynamics:zd_states}. This implies, in
particular, that the total energy ${\cal E}_{\rm zero} := \KEzero +
\PEzero$ is constant along solutions of the swing-phase zero
dynamics. \emph{Note that the energy ${\cal E}_{\rm zero}$ is not the total energy of the initial system evaluated on the zero dynamics.} Indeed, the total energy of the original system is not constant along solutions because actuator power is being injected to create the virtual constraints. One might call ${\cal E}_{\rm zero} $ pseudo-energy; it relation to stability is illustrated in Fig.~\ref{fig:within_stride_ctrl:walking_energytransfer}.

The analysis of periodic orbits of the hybrid zero dynamics is based on the restricted \Poincare\ map. Take the
\Poincare\ section to be $\poincaresection \cap \zdmanifold$ and let
\begin{equation}
  \rho: \poincaresection \cap \zdmanifold \to \poincaresection \cap \zdmanifold
\end{equation}
denote the \Poincare\ (first return) map of the hybrid zero dynamics. Using the fact that the
total energy ${\cal E}_{\rm zero}$ is constant along solutions of the
swing phase zero dynamics, the \Poincare\ map is shown in \cite[pp.~129]{WGCCM07} to
be
\begin{equation}\label{eqn:grizzle:zero_dynamics:poincarezdyn}
  \rho(\zeta^-) = \deltazero^2\,\zeta^- - \PEzero(\theta^-),
\end{equation}
where $\zeta^- := \frac{1}{2} (\sigma^-)^2$, and its domain of
definition is
\begin{equation}\label{eqn:grizzle:zero_dynamics:domain}
  {\cal D}_\zero= \left\{ \zeta^- > 0~\big|~\deltazero^2 \, \zeta^- - \PEzeromax > 0 \right\},
\end{equation}
where
\begin{equation}
  \PEzeromax := \max_{\theta^+ \le \theta \le \theta^-} \PEzero(\theta).
\end{equation}
The domain ${\cal D}_\zero$ is non-empty if, and only if,
$\deltazero^2>0$.
Whenever $\deltazero^2 < 1$, the fixed point of
\eqn{grizzle:zero_dynamics:poincarezdyn},
\begin{equation} \label{eqn:grizzle:zero_dynamics:p_nonlinear_fixed_point}
  \zeta^* := -\frac{\PEzero(\theta^-)}{1-\deltazero^2},
\end{equation}
will be exponentially stable as long as it belongs to ${\cal
D}_\zero$. The conditions for there to exist an exponentially stable
periodic orbit of \eqn{grizzle:zero_dynamics:reduced_hybrid_model} are thus
\begin{subequations}\label{eqn:grizzle:zero_dynamics:zd_conds}
  \begin{gather}\label{eqn:grizzle:zero_dynamics:zd_cond1}
    \frac{\deltazero^2}{1-\deltazero^2}\PEzero(\theta^-) + \PEzeromax < 0\\
    0 < \deltazero^2 < 1.\label{eqn:grizzle:zero_dynamics:zd_cond2}
  \end{gather}
\end{subequations}

\noindent \textbf{Remark:} In the case of a single degree of underactuation,  \eqref{grizzle:zero_dynamics:AngularMomentumTransfer0}-\eqref{grizzle:zero_dynamics:AngularMomentumTransfer2} imply that the \emph{stability of the periodic orbit in the zero dynamics is determined by the direction of the vertical velocity of the center of mass at the end of the step, and is \emph{essentially independent} of the particular choice of virtual constraints used to realize the orbit}. Stated a bit more precisely, different  choices of $\aq$, $\uq$ and $h_d$ in \eqref{eqn:grizzle:FdbkDesignApproach:output_fcn_structure_new_coord}, each with an invertible decoupling matrix, and each defining a virtual constraint that is identically zero along the same periodic orbit, will result in a hybrid zero dynamics with the same stability properties. Other aspects of the transient behavior, such as the basin of attraction or how much control effort is required to return to the periodic orbit, may be different.

\noindent \textbf{Remark:} The motion corresponding to the 3D linear inverted pendulum (3D-LIP), widely used in control of humanoid robots, constrains the center of mass such that $z(t)$ is constant. From \eqref{grizzle:zero_dynamics:AngularMomentumTransfer1}, a similar motion in the zero dynamics would result in $\deltazero=1$, and consequently, would not be exponentially stable.

\subsection{Higher degrees of underactuation}

When two or more degrees of underactuation are considered, the dimension of the zero dynamics is then greater than or equal to four, and the restricted Poincar\'e map \eqref{eqn:grizzle:RestrictedPoincareMap} can no longer be computed in closed form. While its numerical calculation remains a valid means to analyze the stability of a periodic solution,  it is more challenging to extract information for synthesis of a gait. In addition, when there are two or more degrees of underactuation, it has been shown that stability of a periodic motion within the zero dynamics does indeed depend on the choice of the virtual constraints, in contrast to the case of one degree of underactuation analyzed above.  Reference \cite{CHGRSHIH09} gives two choices of virtual constraints that result in the \textit{same periodic solution of the robot}, and yet the periodic orbit is asymptotically stable for one of the choices and unstable for the other; see \cite{BURAHAGRGAGR14} for additional examples. In each of the cited examples, stability was recovered by judiciously modifying a virtual constraint in the ``frontal plane''.

In the case of one degree of underactuation, the freedom coming from the fact that the joint path is controlled but not the time evolution along the path, compensates the underactuation: any path can be followed and the corresponding time evolution along it is unique \cite{CHEV03}, though not freely assignable. Different virtual constraints restraining the robot's configuration to the same path result in the same dynamic behavior. With higher degrees of underactuation, the virtual constraint does not define a unique path in the configuration space, but instead a surface of higher dimension. A periodic trajectory induces a periodic orbit in the state space that belongs to a continuum of virtual constraints defining different surfaces that each include this path. As a consequence, different dynamic behaviors will be obtained.

In a certain sense, the fact that stability depends on the choice of virtual constraints is the more natural situation. The surprise was really that for one degree of underactuation, the periodic orbit itself determines closed-loop stability when designing controllers on the basis of virtual constraints. The next section discusses work in \cite{AKBUGR14} that shows how to systematically search through a family of virtual constraints to find stabilizing solutions, with no restrictions of the degree of underactuation. In parallel to this work, the selection of virtual constraints and switching conditions in order to obtain ``self-synchronization'' of frontal and sagittal plane motions of an underactuated 3D biped is explored in \cite{razavi2014restricted}, along with regulation of the ``pseudo energy'' \eqref{eqn:grizzle:zero_dynamics:lagrangian}.

%% file: sections/7-DesignZeroDynamics.tex
%

This section addresses how to make a concrete choice of the virtual constraints so as to achieve a periodic walking gait that satisfies important physical conditions.

%

\subsection{Starting with an existing periodic trajectory}

Assume that $x(t) = (q(t); \dot{q}(t))$, for $0 \le t < T$ is a periodic solution of \eqref{eqn:grizzle:modeling:model_walk} that has already been computed and verified to meet required conditions on ground reaction forces, actuators bounds, etc. Let
\begin{equation}
\label{eqn:grizzle:design:thetaMonotone}
\theta: \Q \to \mathbb{R}
\end{equation}
be such that its derivative along the periodic trajectory is never zero. $\theta$ is called a \emph{gait timing variable or a gait phasing variable}. Without loss of generality, assume that $\dot{\theta}(t) >0$, and thus $\theta$ is strictly increasing along the periodic solution. Let $(\aq, \uq)$ be a choice of regulated and free variables such that $(\aq, \theta)$ are independent functions in the sense that their Jacobian has full rank. Then \cite[Thm.~6.2, pp.~163]{WGCCM07} gives an explicit construction of a virtual constraint of the form
\begin{equation}
\label{eqn:grizzle:design:SimpleVC}
y=\aq -h_d(\theta)
\end{equation}
that vanishes along the trajectory. Moreover, if the associated decoupling matrix is invertible, then the zero dynamics exists and the periodic orbit belongs to the zero dynamics.

\subsection{Design via parameter optimization}

A periodic solution and the virtual constraints can be designed simultaneously by introducing a finite parametrization of the output
\eqn{grizzle:FdbkDesignApproach:output_fcn_structure_new_coord}. In
particular, the function $h_d$ is constructed from \Bezier\
polynomials, which in turn introduces free parameters $\parama$
into the hybrid zero dynamics \eqn{grizzle:zero_dynamics:reduced_hybrid_model},
\begin{equation}\label{eqn:grizzle:zero_dynamics:reduced_hybrid_model_param}
  \Sigma_{\zero,\parama}: \begin{cases}
    \begin{aligned}
      \dot z & = f_{\zerolbl,\parama}(z), &    z^- & \notin \poincaresection\cap \zdmanifold_\parama\\
      z^+ & = \impactmap(z^-), & z^- & \in \poincaresection\cap
      \zdmanifold_\parama,
    \end{aligned}
    \end{cases}
\end{equation}
through
\begin{equation}
\label{eqn:grizzle:zero_dynamics:VC4Optimization}
  h_\parama(q):=\aq-h_d(\uq,\parama).
\end{equation}

A minimum-energy-like cost criterion
\begin{equation}
\label{eqn:grizzle:zero_dynamics:costFunction}
  J(\parama)=\frac{1}{\mbox{step length}}
  \int_0^{\mbox{\footnotesize step duration}}||u^*_\parama(t)||^2_2dt
\end{equation}
is posed, where $u^*_\parama$ is determined from \eqref{eqn:grizzle:VC:ustarFromZD}, which is much less computationally demanding than \eqref{eqn:grizzle:VC:eq_torque_star}. It is still true that $u^\ast_\parama$
is the unique input to the model \eqn{grizzle:modeling:model_walk} constraining the solution to the zero dynamics surface.  Parameter optimization is then used to (locally) minimize the
cost $J(\parama)$ subject to various equality and inequality
constraints to prescribe walking at a desired average speed,
with the unilateral forces on the support leg lying in the allowed
friction cone, bounds on actuator torques are respected, a minimum swing foot clearance is achieved, and the solution is periodic \cite{WGCCM07,CHGRSHIH09,RAHUAKGR14}.

A solution to the optimization problem results in a set of virtual constraints, namely \eqref{eqn:grizzle:zero_dynamics:VC4Optimization} for a value $\parama^*$, and a periodic orbit of the zero dynamics. In addition, through \eqref{eqn:grizzle:VC:eq_torque}, a controller is produced for the full-dimensional model. Figure \ref{fig:grizzle:modeling:limitcycle} shows a typical stable limit cycle of the closed-loop hybrid system.

References \cite{GrGr2015ACC,HITE12} show how to include common gait perturbations, such as terrain variations, into the optimization. The cost function \eqref{eqn:grizzle:zero_dynamics:costFunction} on the periodic orbit is augmented with terms that account for additional solutions of the model responding to perturbations. In this way, a controller is designed that not only creates a periodic solution on level ground, for example, but also responds appropriately to terrain height changes, slopes, and other perturbations.

\subsection{Systematic search of virtual constraints that guarantee stability}

In the case of one degree of underactuation, one can impose the stability constraint, $0 < \deltazero^2 < 1$, on the optimization problem \eqref{eqn:grizzle:zero_dynamics:costFunction}, though experience shows that it is rarely necessary to do so because the pendulum-like evolution of a biped's center of mass usually results in its velocity pointing downward at the end of the step. On the other hand, when there are two or more degrees of underactuation, the periodic solution resulting from the optimization problem may not be stabilized by the particular set of virtual constraints used in the optimization problem, though the same periodic solution may be stabilized by some other choice of virtual constraints \cite{CHGRSHIH09}. The method of \cite{AKBUGR14,Buss2015} for systematically searching for stabilizing virtual constraints is briefly outlined.


Assume once again that $x(t) = (q(t); \dot{q}(t))$, for $0 \le t < T$ is a periodic solution of \eqref{eqn:grizzle:modeling:model_walk}, and that a choice of gait phasing variable \eqref{eqn:grizzle:design:thetaMonotone} has been made. Let $\aq^\ast(\theta)$ and $\uq^\ast(\theta)$ be the values of the regulated and free variables, respectively, along the periodic orbit, and let $H(\xi)$ be a $k \times (N-k)$ matrix, depending smoothly on a vector of parameters $\xi\in\mathbb{R}^p$. The parameterized virtual constraint
\begin{equation}
\label{eqn:grizzle:zero_dynamics:VC4Search}
y=h(q,\xi):= \aq - \aq^\ast(\theta) + H(\xi)\left(\uq - \uq^\ast(\theta) \right)
\end{equation}
vanishes on the periodic orbit for all values of the parameter vector $\xi$. The matrix $H(\xi)$ forms linear combinations of the free variables, such as roll or yaw, for example. When $H(\xi^*)=0$, \eqref{eqn:grizzle:zero_dynamics:VC4Search} reduces to the nominal virtual constraint resulting from the optimization problem \eqref{eqn:grizzle:zero_dynamics:costFunction}, for example. More genereally, $h(q,\xi)$ could be any smooth function that vanishes on the periodic orbit for all allowed values of the parameters: the linear combinations suggested in \eqref{eqn:grizzle:zero_dynamics:VC4Search} are just one straightforward way to build a family of such functions \cite{CHGRSHIH09}.

Assuming invertibility of the decoupling matrix for a value of $\xi$, say $\xi^*$, the resulting \Poincare\ map will have a fixed point that is independent of the parameters, that is
\begin{equation}
\label{eqn:grizzle:zero_dynamics:VC4SearchPoincare}
x^* = P(x^*,\xi),~\forall \xi \in \mathbb{R}^p~\text{near}~\xi^*.
\end{equation}
 By Taylor's Theorem, it follows that the Jacobian of the \Poincare\ map can be expanded about $\xi^*$ as
\begin{equation}
\label{eqn:grizzle:zero_dynamics:VC4SearchJacobianPoincare}
\frac{\partial P}{\partial x}(x^*,\xi) \approx A_{0} + \sum_{i=1}^{p} A_i \left( \xi_{i} - \xi^*_{i} \right).
\end{equation}
Reference \cite{AKBUGR14} shows how to compute the sensitivity matrices $\left\{A_0, A_1, \ldots, A_p \right\}$, and how to determine if there is a value of $\xi$ near $\xi^*$ resulting in the sum of matrices on the right having eigenvalues in the unit circle. Reference \cite{Buss2015} shows how the same analysis can be performed on the restricted \Poincare\ map, which is perhaps philosophically more satisfying as it is based on the low dimensional pendulum-like dynamics of the underactuated portion of the model.

\subsection{Event-based control}

An alternative method to stabilize a periodic solution is to modify the virtual constraints step to step. To do this, one introduces parameters $\beta$ into the virtual constraints
\begin{equation}\label{eqn:grizzle:EventBasedVCs}
  y = h({q},\beta)
\end{equation}
in such a way that they alter step length, step width, or torso lean angle, for example \cite{WGCCM07,CHGRSHIH09}. In this case, the parameters $\beta$ do modify the periodic orbit, as opposed to the approach of the previous subsection. The \Poincare\ map results in a discrete-time control system
\begin{equation}\label{eqn:grizzle:EventBasedPoincare}
  x_{k+1}=P(x_k,\beta_k).
\end{equation}
If $x^*=P(x^*,\beta^*)$ is a fixed point, then an \textit{event-based control action} of the form
\begin{equation}\label{eqn:grizzle:EventBasedPoincareControl}
  \beta_k = \beta^* + K(x_k - x^*)
\end{equation}
can be designed on the basis of a Jacobian linearization of \eqref{eqn:grizzle:EventBasedPoincare}. The same analysis and design can be carried out with the restricted \Poincare\ map \cite[pp.~107]{WGCCM07}.

A potential limiting factor in using event-based control is that the updates required for stability are only made when the solution crosses a \Poincare\ section, though this can be partly mitigated by using a \Poincare\ section at mid stance, for example, instead of \eqref{eqn:grizzle:modeling:S}. The event-based nature of the updates can induce delays in dealing with perturbations. Because velocity estimates are often ``noisy'', smoothing must be considered when sampling the state vector, which can also induce phase lag.

%% file: sections/8-ExperimentalImplementation.tex
\begin{figure}[t]
      \centering
      \subfloat[simulation\label{fig:grizzle:modeling:limitcycle_sim}]{\includegraphics[scale=1.5]{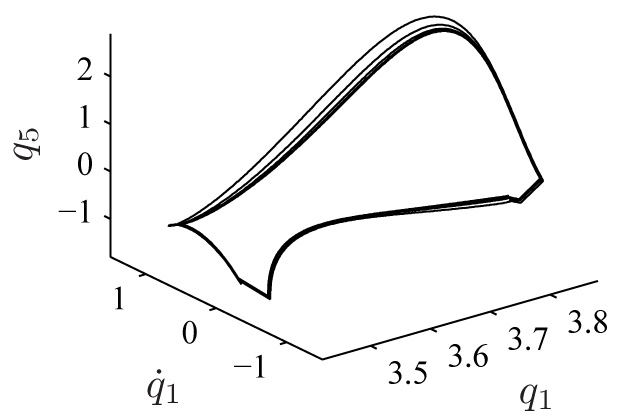}}
      ~~
      \subfloat[experiment\label{fig:grizzle:modeling:limitcycle_exp}]{\includegraphics[scale=1.5]{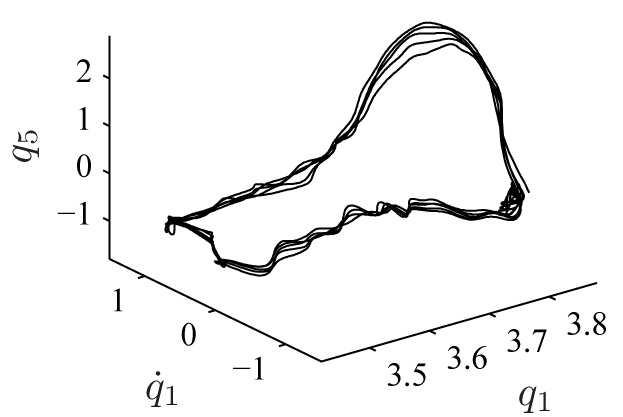}}
    \caption[]{Example limit cycles of a 5-link robot with a zero-dynamics controller.}
    \label{fig:grizzle:modeling:limitcycle}
  \end{figure}

%
%
To date, virtual constraints and hybrid zero dynamics have been implemented on at least nine bipedal robots and three lower-limb prostheses. Figure \ref{fig:grizzle:modeling:limitcycle} shows an example limit cycle. Videos of experiments for the robots Rabbit, MABEL, and MARLO  are available at \cite{GrizWebYouTube2015}. Videos of experiments for the robots AMBER-1, AMBER-2, AMBER-3, NAO, and DURUS  are available at \cite{AmesYouTube2015}. Videos of experiments for the robot ERNIE are available at \cite{SchmiedlerYouTube2015}.  Videos of experiments for the lower-limb prostheses AMPRO and the Vanderbilt Leg are available at \cite{AmesYouTube2015} and \cite{GreggYouTube2015}, respectively. Publications associated with the above robots and prostheses are straightforward to find; they discuss practical aspects of implementing control laws based on virtual constraints.

%% file: Challenges.tex
Many results have not been adequately covered in this overview. The hybrid models of bipedal robots typically have multiple continuous domains \cite{GrChAmSi2014}. The use of  virtual constraints and hybrid zero dynamics in this context can be found in \cite{WGCCM07,SrPaPoGr14,lack2014planar} and references therein. Emphasis on non-trivial double support phases is found in \cite{hereid2014dynamic}. Fully actuated robots are treated in \cite{WGCCM07,WaChTl13,lack2014planar}. Human like motion with foot rolling and double support phases has recently been demonstrated on DURUS \cite{AmesYouTube2015}. The shape of the foot and its deformation can be very important for the walking properties as shown in passive walking. The rolling of a convex foot induces a model of the contact between the ground and the foot that is different from a point-foot contact, but it still involves underactuation \cite{martin2014effects}.

The analysis procedures and control designs presented in the chapter have focused on periodic locomotion, a form of steady-state behavior in a hybrid model of walking. It is important to move beyond this assumption and address aperiodic or transient motions. There is room for improved notions of stability of aperiodic walking gaits. Two cases where aperiodic gaits arise naturally are walking on uneven ground and maneuvering a biped around obstacles. Composition of motion primitives as a means to handle aperiodic (uneven) terrain is featured in \cite{yang2009framework,powell2012motion,PaRaGr13}. The introduction of elasticy in legged robots can improve robustness and adaptation to uneven terrain. Series elastic actuators are treated in \cite{POGR09,SrPaPoGr2011,SrPaPoGr14,RAHUAKGR14,powellhierarchical}.

How to choose the virtual constraints is an important question. In this chapter, parametric functions (or splines) were suggested, with the unknown parameters selected to optimize a given criterion. The choice of what to control has been partially addressed in \cite{hamed2015exponentially}. The optimization criterion can be selected to account for uncertainty in the hybrid model. Terrain variation is addressed in \cite{GrGr2015ACC}. Nonholonomic virtual constraints have just been introduced in \cite{GrGr2015CDC}; they allow swing foot placement to be planned as a function of velocity. A systematic means to get started with virtual constraints is provided in Appendix A of \cite{WGCCM07}.

Characterizing the domain of attraction of a periodic walking gait for a bipedal robot model is still in its infancy. Analysis via SOS (Sums of Squares) is investigated in \cite{manchester2011transverse}.

Many other interesting questions arise, ranging from reflex actions to enhance stability under large perturbations, to bipedal robot safety when operating around humans, manipulation of objects, navigation, etc.

%
%
%
%
%
%

%% file: HZD_Handbook.bbl
\begin{thebibliography}{10}

\bibitem{AKBUGR14}
K.~Akbari~Hamed, B.~G. Buss, and J.~W. Grizzle.
\newblock Continuous-time controllers for stabilizing periodic orbits of hybrid
  systems: Application to an underactuated {3D} bipedal robot.
\newblock In {\em Decision and Control (CDC), 2014 IEEE 53rd Annual Conference
  on}, pages 1507--1513, Dec 2014.

\bibitem{AKGR14}
K.~{Akbari Hamed} and J.W Grizzle.
\newblock Event-based stabilization of periodic orbits for underactuated 3-d
  bipedal robots with left-right symmetry.
\newblock {\em Robotics, IEEE Transactions on}, 30(2):365--381, 2014.

\bibitem{ames2014human}
A.~D. Ames.
\newblock Human-inspired control of bipedal walking robots.
\newblock {\em Automatic Control, IEEE Transactions on}, 59(5):1115--1130,
  2014.

\bibitem{AmesYouTube2015}
A.~D. Ames.
\newblock {AMBER-Lab}.
\newblock {\em \url{https://www.youtube.com/user/ProfAmes}}, November 2016.

\bibitem{ames2014rapidly}
A.D. Ames, K.~Galloway, K.~Sreenath, and {J.W.} Grizzle.
\newblock Rapidly exponentially stabilizing control lyapunov functions and
  hybrid zero dynamics.
\newblock {\em IEEE Trans. Automatic Control}, 59(4):876--891, 2014.

\bibitem{BURAHAGRGAGR14}
B.G. Buss, A.~Ramezani, K.~{Akbari Hamed}, B.A. Griffin, K.S. Galloway, and
  J.W. Grizzle.
\newblock Preliminary walking experiments with underactuated {3D} bipedal robot
  {MARLO}.
\newblock In {\em Intelligent Robots and Systems (IROS 2014), 2014 IEEE/RSJ
  International Conference on}, pages 2529--2536, Sept 2014.

\bibitem{Buss2015}
Brian~G. Buss.
\newblock {\em Systematic Controller Design for Dynamic 3D Bipedal Robot
  Walking}.
\newblock PhD thesis, University of Michigan, Ann Arbor, MI, USA, May 2015.

\bibitem{BYRNESC91}
C.~Byrnes and A.~Isidori.
\newblock Asymptotic stabilization of nonlinear minimum phase systems.
\newblock {\em IEEE Trans. Autom. Contr.}, 376:1122--37, 1991.

\bibitem{CHEV03}
C.~Chevallereau.
\newblock Time scaling control for an underactuated biped robot.
\newblock {\em IEEE Transactions on Robotics and Automation}, 19(2):362--368,
  April 2003.

\bibitem{CHetal03}
C.~Chevallereau, G.~Abba, Y.~Aoustin, F.~Plestan, E.~R. Westervelt, C.~Canudas,
  and J.~W. Grizzle.
\newblock {RABBIT:} a testbed for advanced control theory.
\newblock {\em IEEE Contr. Syst. Mag.}, 23(5):57--79, October 2003.

\bibitem{ChDjGr08}
C.~Chevallereau, D.~Djoudi, and Journal = TRO Number = 2 Pages = {390--401}
  Title = {Stable Bipedal Walking with Foot Rotation Through Direct Regulation
  of the Zero Moment Point} Volume = 25 Year =~2008 Grizzle, J.~W.

\bibitem{CHGRSHIH09}
C.~Chevallereau, J.~W. Grizzle, and {C.-L.} Shih.
\newblock Asymptotically stable walking of a five-link underactuated {3D}
  bipedal robot.
\newblock {\em IEEE Trans. Robot.}, 25(1):37--50, 2009.

\bibitem{HITE12}
Hongkai Dai and R.~Tedrake.
\newblock Optimizing robust limit cycles for legged locomotion on unknown
  terrain.
\newblock In {\em Decision and Control (CDC), 2012 IEEE 51st Annual Conference
  on}, pages 1207--1213, 2012.

\bibitem{GreggYouTube2015}
R.D. Gregg.
\newblock High-performance control of a powered transfemoral prosthesis with
  amputee subjects.
\newblock {\em \url{https://www.youtube.com/watch?v=sl1IXs0j4Ww}}, November
  2016.

\bibitem{gregg2014virtual}
R.D. Gregg, T.~Lenzi, L.J. Hargrove, and J.W. Sensinger.
\newblock Virtual constraint control of a powered prosthetic leg: From
  simulation to experiments with transfemoral amputees.
\newblock {\em IEEE Transactions on Robotics}, 2014.

\bibitem{gregg2014evidence}
R.D. Gregg, E.J. Rouse, L.J. Hargrove, and J.W. Sensinger.
\newblock Evidence for a time-invariant phase variable in human ankle control.
\newblock {\em {PlOS ONE}}, 9(2):e89163, 2014.

\bibitem{GrGr2015CDC}
B.~Griffin and J.W. Grizzle.
\newblock Nonholonomic virtual constraints for dynamic walking.
\newblock In {\em Preprint submitted to IEEE Conference on Decision and
  Control}, 2015.

\bibitem{GrGr2015ACC}
B.~Griffin and J.W. Grizzle.
\newblock Walking gait optimization for accommodation of unknown terrain height
  variations.
\newblock In {\em American Control Conference}, 2015.

\bibitem{GrizWebYouTube2015}
J.~W. Grizzle.
\newblock {Dynamic Leg Locomotion}.
\newblock {\em \url{www.youtube.com/user/DynamicLegLocomotion}}, November 2016.

\bibitem{GRABPL01}
J.~W. Grizzle, G.~Abba, and F.~Plestan.
\newblock Asymptotically stable walking for biped robots: Analysis via systems
  with impulse effects.
\newblock {\em IEEE Trans. Autom. Contr.}, 46:51--64, January 2001.

\bibitem{GrChAmSi2014}
J.W. Grizzle, C.~Chevallereau, A.~Ames, and R.~Sinnet.
\newblock Models, feedback control, and open problems of {3D} bipedal robotic
  walking.
\newblock {\em Automatica}, 50(8):1955--1988, 2014.

\bibitem{hamed2015exponentially}
Kaveh~Akbari Hamed, Brian~G Buss, and Jessy~W Grizzle.
\newblock Exponentially stabilizing continuous-time controllers for periodic
  orbits of hybrid systems: {A}pplication to bipedal locomotion with ground
  height variations.
\newblock {\em The International Journal of Robotics Research}, 35(8):977--999,
  July 2016.

\bibitem{hereid2014dynamic}
A.~Hereid, S.~Kolathaya, M.S. Jones, J.~Van~Why, J.W. Hurst, and A.D. Ames.
\newblock Dynamic multi-domain bipedal walking with {ATRIAS} through {SLIP}
  based human-inspired control.
\newblock In {\em Proceedings of the 17th international conference on Hybrid
  systems: computation and control}, pages 263--272. ACM, 2014.

\bibitem{HoReZhPaAm2015}
J.~Horn, J.~Reher, Zhao, V.~Paredes, and A.D. Ames.
\newblock {AMPRO}: Translating robotic locomotion to a powered transfemoral
  prosthesis.
\newblock In {\em International Conference on Robotics and Automation, ICRA},
  May 2015.

\bibitem{HUMA94}
Y.~H\"{u}rm\"{u}zl\"{u} and D.~B. Marghitu.
\newblock Rigid body collisions of planar kinematic chains with multiple
  contact points.
\newblock {\em Int. J. Robot. Res.}, 13(1):82--92, 1994.

\bibitem{lack2014planar}
J.~Lack, M.J. Powell, and A.D. Ames.
\newblock Planar multi-contact bipedal walking using hybrid zero dynamics.
\newblock In {\em Robotics and Automation (ICRA), 2014 IEEE International
  Conference on}, pages 2582--2588. IEEE, 2014.

\bibitem{manchester2011transverse}
Ian~R Manchester.
\newblock Transverse dynamics and regions of stability for nonlinear hybrid
  limit cycles.
\newblock {\em IFAC Proceedings Volumes}, 44(1):6285--6290, 2011.

\bibitem{martin2014effects}
Anne~E Martin, David~C Post, and James~P Schmiedeler.
\newblock The effects of foot geometric properties on the gait of planar bipeds
  walking under {HZD}-based control.
\newblock {\em The International Journal of Robotics Research},
  33(12):1530--1543, 2014.

\bibitem{MORRISB05}
B.~Morris and J.~W. Grizzle.
\newblock A restricted {P}oincar\'e map for determining exponentially stable
  periodic orbits in systems with impulse effects: Application to bipedal
  robots.
\newblock In {\em Proc. of the 2005 IEEE International Conference on Decision
  and Control European Control Conference, Seville, Spain}, pages 4199--206,
  2005.

\bibitem{MOGR08}
B.~Morris and J.~W. Grizzle.
\newblock Hybrid invariant manifolds in systems with impulse effects with
  application to periodic locomotion in bipedal robots.
\newblock {\em IEEE Trans. Autom. Contr.}, 54(8):1751--1764, 2009.

\bibitem{PaRaGr13}
{H.-W.} Park, A.~Ramezani, and J.~W. Grizzle.
\newblock A finite-state machine for accommodating unexpected large ground
  height variations in bipedal robot walking.
\newblock {\em IEEE Trans. Robot.}, 29(29):331--345, 2013.

\bibitem{POGR09}
I.~Poulakakis and J.~W. Grizzle.
\newblock The spring loaded inverted pendulum as the hybrid zero dynamics of an
  asymmetric hopper.
\newblock {\em IEEE Trans. Autom. Contr.}, 54(8):1779--1793, 2009.

\bibitem{powellhierarchical}
M.J. Powell and A.D. Ames.
\newblock Hierarchical control of series elastic actuators through control
  lyapunov functions.

\bibitem{powell2012motion}
M.J. Powell, H.~Zhao, and A.D. Ames.
\newblock Motion primitives for human-inspired bipedal robotic locomotion:
  walking and stair climbing.
\newblock In {\em Robotics and Automation (ICRA), 2012 IEEE International
  Conference on}, pages 543--549, 2012.

\bibitem{RAHUAKGR14}
A~Ramezani, J.~W. Hurst, K.~{Akbari Hamed}, and J.~W. Grizzle.
\newblock Performance analysis and feedback control of {ATRIAS}, a
  three-dimensional bipedal robot.
\newblock {\em J. Dynamic Syst., Measurement, Contr.},
  136(2):0210112--1--0210112--12, March 2014.

\bibitem{razavi2014restricted}
H.~Razavi, A.M. Bloch, C.~Chevallereau, and J.W. Grizzle.
\newblock Restricted discrete invariance and self-synchronization for stable
  walking of bipedal robots.
\newblock In {\em American Control Conference}, 2015.

\bibitem{sadati2012hybrid}
N.~Sadati, G.~A. Dumont, K.~A. Hamed, and W.~A. Gruver.
\newblock {\em Hybrid Control and Motion Planning of Dynamical Legged
  Locomotion}.
\newblock IEEE Press Series on Systems Science and Engineering. Wiley, 2012.

\bibitem{SchmiedlerYouTube2015}
J.~Schmiedeler.
\newblock {ERNIE} robot walking with different feet.
\newblock {\em \url{https://www.youtube.com/watch?v=T2x3VvPaacA}}, November
  2016.

\bibitem{ShGrCh12}
{C-L.} Shih, {J.W.} Grizzle, and {C.} Chevallereau.
\newblock From stable walking to steering of a {3D} bipedal robot with passive
  point feet.
\newblock {\em Robotica}, 30(7):1119--1130, December 2012.

\bibitem{shiriaev2010transverse}
A.S. Shiriaev, L.B. Freidovich, and S.V. Gusev.
\newblock Transverse linearization for controlled mechanical systems with
  several passive degrees of freedom.
\newblock {\em Automatic Control, IEEE Transactions on}, 55(4):893--906, 2010.

\bibitem{shiriaev2008can}
A.S. Shiriaev, L.B. Freidovich, and I.R. Manchester.
\newblock Can we make a robot ballerina perform a pirouette? orbital
  stabilization of periodic motions of underactuated mechanical systems.
\newblock {\em Annual Reviews in Control}, 32(2):200--211, 2008.

\bibitem{SrPaPoGr2011}
K.~Sreenath, {H.-W.} Park, I.~Poulakakis, and J.~W. Grizzle.
\newblock A compliant hybrid zero dynamics controller for stable, efficient and
  fast bipedal walking on {MABEL}.
\newblock {\em Int. J. Robot. Res.}, 30(9):1170--1193, 2011.

\bibitem{SrPaPoGr14}
K.~Sreenath, {H.-W.} Park, I.~Poulakakis, and J.~W. Grizzle.
\newblock Embedding active force control within the compliant hybrid zero
  dynamics to achieve stable, fast running on {MABEL}.
\newblock {\em Int. J. Robot. Res.}, 33:988--1005, 2014.

\bibitem{Zutven:PHD}
P.~{van Zutven}.
\newblock {\em Control and Identification of Bipedal Humanoid Robots: Stability
  Analysis and Experiments}.
\newblock PhD thesis, University of Technology, Eindhoven, The Netherlands,
  2014.

\bibitem{VUBOSUST90}
M.~Vukobratovi{\'c}, B.~Borovac, D.~Surla, and D.~Stokic.
\newblock {\em Biped Locomotion}.
\newblock Springer-Verlag, Berlin, 1990.

\bibitem{WaChTl13}
T.~Wang, C.~Chevallereau, and D.~Tlalolini.
\newblock Stable walking control of a {3D} biped robot with foot rotation.
\newblock {\em Robotica}, FirstView:1--20, March 2014.

\bibitem{WEBUGR04}
E.~R. Westervelt, G.~Buche, and J.~W. Grizzle.
\newblock Experimental validation of a framework for the design of controllers
  that induce stable walking in planar bipeds.
\newblock {\em Int. J. Robot. Res.}, 24(6):559--582, 2004.

\bibitem{WGCCM07}
E.~R. Westervelt, J.~W. Grizzle, C.~Chevallereau, {J.-H.} Choi, and B.~Morris.
\newblock {\em Feedback Control of Dynamic Bipedal Robot Locomotion}.
\newblock Boca Raton, 2007.

\bibitem{WEGRKO03}
E.~R. Westervelt, J.~W. Grizzle, and D.~E. Koditschek.
\newblock Hybrid zero dynamics of planar biped walkers.
\newblock {\em IEEE Trans. Autom. Contr.}, 48(1):42--56, January 2003.

\bibitem{yang2009framework}
T.~Yang, E.~R. Westervelt, A.~Serrani, and J.~P. Schmiedeler.
\newblock A framework for the control of stable aperiodic walking in
  underactuated planar bipeds.
\newblock {\em Auton. Robots}, 27(3):277--290, 2009.

\bibitem{zhao2014human}
{H.-H.} Zhao, {W.-L.} Ma, {A. D.} Ames, and {M. B.} Zeagler.
\newblock Human-inspired multi-contact locomotion with {AMBER2}.
\newblock In {\em Cyber-Physical Systems (ICCPS), 2014 ACM/IEEE International
  Conference on}, pages 199--210, 2014.

\end{thebibliography}
